\documentclass[a4paper,11pt]{article}
\usepackage[english]{babel}
\usepackage[margin=3.0cm]{geometry}
\usepackage{amsmath}
\usepackage{amsthm}
\usepackage{amssymb}
\usepackage[utf8]{inputenc}
\usepackage{hyperref}
\usepackage{cleveref}
\usepackage{autonum}
\usepackage{booktabs}
\hypersetup{	unicode}
\usepackage{graphicx, color}
\graphicspath{{fig/}}
\usepackage{algorithm, algpseudocode} 
\usepackage{mathrsfs} 
\usepackage{authblk}
\usepackage{tikz}
\usetikzlibrary{positioning,calc,arrows.meta,fit,shapes.geometric,backgrounds}
\usepackage{adjustbox}
\usepackage{tabularx}
\usepackage{subcaption}

\newtheorem{remark}{\bf Remark}[section]

\providecommand{\keywords}[1]{\textbf{\textit{Keywords: }} #1}

\title{Asymptotic-Preserving Neural Networks for Pressure Reconstruction in Multiscale Blood Flow Modeling}

\author{Giulia Bertaglia}
\author{Raffaella Fiamma Cabini \footnote{Corresponding author, e-mail: raffaellafiamma.cabini@unife.it}}

\affil{Department of Environmental and Prevention Sciences, University of Ferrara, Corso Ercole I d’Este 32, 44121, Ferrara, Italy}

\begin{document}
\date{November 17, 2026}
\maketitle
%%==================================%%
%% ABSTRACT %%
%%==================================%%
\begin{abstract}
Mathematical models and numerical simulations offer a non-invasive way to explore cardiovascular phenomena, providing access to quantities that cannot be measured directly. In this study, we start with a one-dimensional multiscale blood flow model that describes the viscoelastic properties of arterial walls, and we focus on improving its practical applicability by addressing a major challenge: determining, in a reliable way, the viscoelastic parameters that control how arteries deform under pulsatile pressure. To achieve this, we employ Asymptotic-Preserving Neural Networks that embed the governing physical principles of the multiscale blood flow model within the learning procedure. This framework allows us to infer the viscoelastic parameters while simultaneously reconstructing the time-dependent evolution of the state variables of blood vessels and remaining consistent with all the multiscale regimes of the system. With this approach, pressure waveforms are estimated from readily accessible patient-specific data, i.e., cross-sectional area and velocity measurements from Doppler ultrasound, in vascular segments where direct pressure measurements are not available. Different numerical simulations, conducted in both synthetic and patient-specific scenarios, show the effectiveness of the proposed methodology.
\end{abstract}


\tableofcontents

%%==================================%%
%% INTRODUCTION %%
%%==================================%%
\section{Introduction}
The human cardiovascular system is a complex and dynamic network that transports blood throughout the body, delivering oxygen and nutrients to tissues while removing metabolic waste products~\cite{levick2013introduction, caro2012mechanics}. Its function emerges from the coordinated interaction between the heart, which generates pulsatile flow, and the vascular network, whose vessels adapt and respond to these flow patterns. The interaction between blood flow and the mechanical properties of the vessel walls produces pulsatile waves in vessel cross-sectional area, blood velocity, and intravascular pressure that propagate along the network. These hemodynamic signals reflect cardiac activity while also encoding information about the mechanical state and structure of the vessels. For example, blood pressure is a key risk factor that affects arterial integrity and contributes to cardiovascular events~\cite{safar2003current, safar2001systolic, alastruey2012arterial}. Thus, monitoring hemodynamic variables is essential for understanding cardiovascular physiology and developing diagnostic tools. However, not all hemodynamic quantities can be easily measured non-invasively. While vessel cross-sectional area and blood velocity can be obtained non-invasively using Doppler ultrasound or magnetic resonance imaging, measuring pressure in non-superficial vessels remains challenging and typically requires invasive procedures~\cite{meidert2018techniques, scheer2002clinical}.

In this context, mathematical modeling and numerical simulations have emerged as powerful tools for studying cardiovascular dynamics and estimating these quantities non-invasively~\cite{quarteroni2004mathematical, formaggia2010cardiovascular, alastruey2007modelling,pfaller2022}. Among the available approaches, one-dimensional (1D) models of blood flow represent an effective compromise between computational efficiency and physical fidelity. By describing the evolution of flow variables along deformable vessels, these models capture the main features of pulse wave propagation while remaining computationally tractable for large vascular networks~\cite{xiao2014systematic}. However, in the traditional formulation the mechanical response of the vessel wall is described using a purely elastic relation between pressure and cross-sectional area. This assumption provides only a first-order approximation of vessel mechanics and neglects the viscoelastic behavior of real vessel walls~\cite{valdez2008analysis, alastruey2012physical,coccarelli2021}, which can lead to an erroneous estimation of pressure peaks~\cite{alastruey2011pulse, holenstein1980viscoelastic}.

To address this limitation, the 1D viscoelastic blood flow model proposed in~\cite{bertaglia2020modeling, bertaglia2020computational, piccioli2021, bertaglia2023multiscale} adopts a Standard Linear Solid (SLS) constitutive law to represent the mechanical response of the vessel wall, which accounts for both the instantaneous elastic response and the time-dependent viscous relaxation. This formulation enables a more realistic representation of fluid-structure interactions and provides an improved description of hemodynamic variables along the vascular network. However, its application requires detailed knowledge of the mechanical properties of the vessels, particularly the viscoelastic coefficients. These parameters cannot be measured directly in vivo and can be made available only through indirect estimates~\cite{bertaglia2020computational}. Moreover, the viscoelastic model exhibits a multiscale behavior which admits different asymptotic limits depending on the values of the viscoelastic parameters, which poses additional challenges for numerical methods used to approximate its solution~\cite{bertaglia2023multiscale}.

In recent years, Physics-Informed Neural Networks (PINNs) have emerged as a powerful framework for approximating solutions of systems of partial differential equations (PDEs)~\cite{karniadakis2021physics, raissi2019physics}. In this approach, the neural network learns to match the observed data while maintaining consistency with the governing equations by minimizing a loss function that accounts for both data mismatch and PDE residuals. This allows the network to reconstruct solutions even when observations are sparse, noisy, or incomplete, while preserving physical consistency. Standard PINNs have been successfully applied to a wide range of problems~\cite{barzaghi2024myo}, including blood flow modeling \cite{kissas2020,garay2024,aghaee2025,orera2026, orera2025}; however, when the governing equations exhibit a multiscale behavior, scaling parameters may induce different asymptotic regimes, and conventional PINN formulations may fail to recover the correct limiting behavior of the system~\cite{jin2023asymptotic,bertaglia2021asymptotic}.

To overcome this limitation, Asymptotic-Preserving Neural Networks (APNNs) have recently been introduced~\cite{jin2023asymptotic, bertaglia2021asymptotic, bertaglia2022asymptotic}. APNNs extend the PINN framework by embedding asymptotic-preserving (AP) properties into the training process, ensuring that network predictions remain consistent with the asymptotic limits of the underlying model across all relevant parameter regimes \cite{jin2022}. This property makes APNNs essential when dealing with multiscale problems, where different physical regimes may arise depending on the values of the governing parameters.

In this work, we apply the APNN framework to the viscoelastic blood flow model proposed in~\cite{bertaglia2020modeling, bertaglia2023multiscale}. By combining easily measurable hemodynamic data (specifically, vessel cross-sectional area and blood velocity) with the physics-based model, the network reconstructs pressure waveforms in the vascular segment of interest. At the same time, the viscoelastic coefficients, which cannot be measured directly in vivo, are treated as learnable inverse parameters and inferred automatically during training. This approach also enables the estimation of hemodynamic variables, including pressure, at vessel locations where direct measurements are unavailable. The AP formulation embedded in the construction of the neural network ensures that predictions remain physically consistent across all relevant regimes, yielding accurate, interpretable reconstructions of the hemodynamic fields.

The rest of this paper is organized as follows. Section~\ref{sec2} introduces the 1D viscoelastic blood flow model and its multiscale asymptotic limits. Section~\ref{sec3} describes the APNN architecture and the theoretical basis of the AP property. Section~\ref{sec4} presents the application of APNNs to the blood flow model. Section~\ref{sec_dat} details the synthetic and in vivo datasets, the applied pre-processing procedures, and the network design and training strategy. Section~\ref{sec5} presents and discusses the numerical results. Finally, Section~\ref{sec6} summarizes the conclusions and outlines directions for future research.

\section{Materials and Methods}
\subsection{Multiscale viscoelastic blood flow model}\label{sec2}

The blood flow model adopted in this work is the 1D viscoelastic formulation introduced by~\cite{bertaglia2020modeling, bertaglia2020computational, piccioli2021, bertaglia2023multiscale}, to which the reader is referred for a detailed description.

Following the standard approach, the 1D mathematical model for blood flow in medium and large arteries is obtained by integrating the three-dimensional incompressible Navier-Stokes equations over the vessel cross-section, under the assumption of axial symmetry of the vessel and of the flow~\cite{formaggia2010cardiovascular}. This procedure leads to a system of balance laws expressing the conservation of mass and momentum.

To close the system, a constitutive relation, referred to as a tube law, is required to relate the internal pressure to the cross-sectional area of the vessel. In many classical 1D blood flow models, this relation is assumed to be purely elastic, providing a reasonable first approximation of vessel behavior~\cite{muller2014global, liang2009biomechanical, boileau2015benchmark, willemet2015arterial}. However, the vessel wall is a complex multi-layer viscoelastic structure whose deformation under blood pressure involves energy dissipation, resulting in hysteresis in the pressure-area relationship. Modeling the blood-wall interaction as purely elastic neglects this dissipation and can lead to overestimation of pressure peaks.  To account for these effects, a viscoelastic constitutive law based on the SLS model was incorporated into the 1D blood flow model, as proposed in~\cite{bertaglia2020computational, bertaglia2020modeling, bertaglia2023multiscale}. Although vessel walls exhibit nonlinear viscoelastic behavior, the SLS-based formulation represents an effective compromise between model complexity and physiological fidelity, as previously demonstrated in 1D cardiovascular modeling studies, including applications involving hypertensive conditions~\cite{bertaglia2020computational, alastruey2011pulse, celant2023modeling, alastruey2012arterial}.

%\begin{figure}
%    \centering
%    \includegraphics[width=0.75\textwidth]{fig/vaso.png}
%\caption{\textcolor{black}{Questa figura si potrebbe togliere} \textcolor{blue}{Sì, io la toglierei} Schematic representation of the one-dimensional blood vessel model.
%The flow is described along the axial coordinate $x$ and time $t$, with cross-sectional area $A(x,t)$, averaged velocity $u(x,t)$, and averaged internal pressure $P(x,t)$. The mechanical response of the vessel wall is characterized by the instantaneous and asymptotic Young moduli, $E_0(x)$ and $E_\infty(x)$, and the wall viscosity $\eta$. The vessel geometry is defined by length $L_0$ and wall thickness $h_0$.}
%\label{fig:vess}
%\end{figure}

Denoting by $A(x,t)$ the cross-sectional area of the vessel, $u(x,t)$ the averaged fluid velocity, $p(x,t)$ the averaged fluid pressure, and $x$ and $t$ the space and time respectively%(see Figure~\ref{fig:vess})
, the resulting 1D viscoelastic blood flow model constitutes a system of PDEs and reads
\begin{equation}
\begin{aligned}
    & \frac{\partial A}{\partial t} + \frac{\partial (A u)}{\partial x} = 0, \\
    & \frac{\partial (A u)}{\partial t} + \frac{\partial (A u^2)}{\partial x} + \frac{A}{\rho}\,\frac{\partial p}{\partial x} = 0, \\
    & \frac{\partial p}{\partial t} + E_0 G(A)\,\frac{\partial (A u)}{\partial x} = -{1\over\tau_r}\bigl(p - F(A)\bigr).
\end{aligned}
\label{eq:visc_model}
\end{equation}
Here, $\rho$ denotes the fluid density, $E_0(x)$ is the instantaneous Young modulus, and $\tau_r(x)$ is the relaxation time associated with the viscoelastic response of the vessel wall, defined by the SLS model, as 
\begin{equation}
    \tau_r = \frac{\eta(E_0-E_\infty)}{E_0^2},
\label{eq:taur}
\end{equation}
where $E_\infty(x)$ and $\eta(x)$ are the asymptotic Young modulus and the wall viscosity coefficient, respectively. \textcolor{black}{Notice that, in the applications presented in the following of this work, each case concerns a single arterial segment, and all mechanical parameters are assumed to be spatially constant within that segment. The momentum equation also neglects viscous friction in the blood, whose effect on the solutions considered here is negligible with respect to that of viscoelastic damping in the vessel wall.}
In system~\eqref{eq:visc_model}, the term $E_0 G(A)$ represents the elastic contribution of the vessel wall, where
\begin{equation}
G(A) = \frac{1}{W A}\left(m \alpha^m - n \alpha^n\right),
\end{equation}
and $\alpha = A / A_0$ denotes the non-dimensional cross-sectional area scaled by the equilibrium area $A_0(x)$. The parameters $m$ and $n$ characterize the mechanical behavior of the vessel wall and depend on whether the vessel is an artery or a vein. The coefficient $W(x)$ depends on the wall thickness $h_0$ (assumed constant) and the equilibrium inner radius $R_0(x)$, and takes different forms for arteries and veins. Specifically, these parameters are given by
\begin{equation} 
\begin{aligned} 
W &= \begin{cases} 
R_0/h_0 & \text{if artery} \\ 
12 R_0^3/h_0^3 & \text{if vein} 
\end{cases} \\[1ex]
m &= \begin{cases} 
1/2 & \text{if artery} \\ 
10 & \text{if vein} 
\end{cases} \\[1ex]
n &= \begin{cases} 
0 & \text{if artery} \\
-3/2 & \text{if vein} 
\end{cases}
\end{aligned} 
\end{equation}
The source term in the third equation of~\eqref{eq:visc_model} accounts for viscous dissipation in the vessel wall, where
\begin{equation}
F(A) = p_0 + \frac{E_\infty}{W}\left(\alpha^m - \alpha^n\right).
\end{equation}
and $p_0(x)$ is the equilibrium pressure.

Following the SLS constitutive law, it is possible to define the relaxation function that describes the time-dependent mechanical behavior of the vessel wall stiffness, $E(t)$:
\begin{equation}
E(t) = E_0 e^{-{t\over\tau_r}} + E_\infty\left(1-e^{-{t\over\tau_r}}\right)
\label{eq:relaxation}
\end{equation}
which starts with the instantaneous value $E_0$ and decays exponentially to the asymptotic value $E_\infty$. The relaxation time $\tau_r$ controls how rapidly the material transitions from the instantaneous to the asymptotic response. 

Introducing the vector of conserved variables
\begin{equation}
\mathbf{Q} =
\begin{pmatrix}
A \\
A u \\
p
\end{pmatrix},
\end{equation}
the model~\eqref{eq:visc_model} can be written in compact form as
\begin{equation}
\partial_t \mathbf{Q} + \partial_x \mathbf{f}(\mathbf{Q}) + \mathbf{B}(\mathbf{Q}) \partial_x \mathbf{Q} = \mathbf{S}(\mathbf{Q}),
\end{equation}
where the flux vector, the non-conservative matrix, and the source term are
\begin{equation}
\begin{aligned}
\mathbf{f}(\mathbf{Q}) &= \begin{pmatrix} A u \\ A u^2 \\ 0 \end{pmatrix}, \\[2ex]
\mathbf{B}(\mathbf{Q}) &= \begin{pmatrix} 0 & 0 & 0 \\ 0 & 0 & A/\rho \\ 0 & E_0 G(A) & 0 \end{pmatrix}, \\[2ex]
\mathbf{S}(\mathbf{Q}) &= \begin{pmatrix} 0 \\ 0 \\ -\frac{1}{\tau_r}(p - F(A)) \end{pmatrix}.
\end{aligned}
\end{equation}
Defining the \textcolor{black}{quasi-linear matrix} %extended Jacobian
$\mathbf{J}(\mathbf{Q}) = {\partial \mathbf{f}}/{\partial \mathbf{Q}} + \mathbf{B}(\mathbf{Q})$,
the system can be equivalently expressed in quasi-linear form:
\begin{equation}
\partial_t \mathbf{Q} + \mathbf{J}(\mathbf{Q}) \partial_x \mathbf{Q} = \mathbf{S}(\mathbf{Q}).
\end{equation}
Because of the presence of the relaxation source term and since $\mathbf{J}(\mathbf{Q})$ is diagonalizable with real eigenvalues ($\lambda_1 = u - c$, $\lambda_2 = 0$, $\lambda_3 = u + c$) and a complete set of linearly independent eigenvectors, model~\eqref{eq:visc_model} results in a hyperbolic relaxation system~\cite{bertaglia2020modeling}. Here, $c$ is the wave speed, defined by
\begin{equation}
c = \sqrt{\frac{AE_0}{\rho}G(A)}.
\label{eq:cele}
\end{equation}

\subsubsection{Viscoelastic parameters calibration}\label{sec:calibration}
Once the viscoelastic parameters $E_0$, $E_\infty$, and $\tau_r$ are calibrated, model~\eqref{eq:visc_model} reproduces the coupled dynamics of $A$, $u$, and $p$. In practice, however, these parameters are difficult to measure directly with high accuracy~\cite{bertaglia2020computational}.

In the purely elastic formulation, the single Young modulus $E = E_0 = E_{\infty}$ is typically calibrated from a reference wave speed $c$, obtained from experimental measurements or the literature, by inverting relation~\eqref{eq:cele}.

In the viscoelastic SLS formulation, the wall elasticity is described by two moduli: the instantaneous modulus $E_0$ and the asymptotic modulus $E_\infty$. To ensure consistency with the elastic model in the long-term limit, $E_\infty$ is calibrated accounting for the wave speed $c$. %The explicit expression of $E_\infty$ differs between arteries and veins; in this work, only the arterial expression is reported, as the subsequent applications focus on arterial vessels. 
Thus, assuming a reference (diastolic) state in which the vessel area $A$ coincides with the equilibrium area $A_0$, $E_\infty$ can be expressed as~\cite{alastruey2012arterial, bertaglia2020modeling}
\begin{equation}
E_\infty =
\begin{cases} 
\displaystyle \frac{2 R_0 \rho c^2}{h_0} & \text{if artery}\\[2mm]
\displaystyle \frac{24 R_0^3 \rho c^2}{23 h_0^3} & \text{if vein}
\end{cases}.
\end{equation}
%where $R_0$ is the equilibrium (diastolic) inner radius of the vessel~\cite{alastruey2012arterial, bertaglia2020modeling}.

The calibration of the remaining two mechanical coefficients, $E_0$ and $\tau_r$, is much more challenging. The reader is referred to~\cite{bertaglia2020computational} for a description of an alternative estimation strategy for these parameters.
In the present study, we consider that only $E_\infty$ can be directly calibrated, following the above mentioned procedure, whereas $E_0$ and $\tau_r$ are determined automatically through the computational procedure detailed in Section \ref{sec4}. 

\subsubsection{Multiscale behavior}
\begin{figure*}
\centering
%\begin{adjustbox}{width=0.75\textwidth}
    \begin{tikzpicture}[
  node distance=1.8cm,
  box/.style={
      draw=black, rectangle, rounded corners, 
      minimum width=4.5cm, minimum height=2cm, 
      text width=4.2cm, align=center, fill=blue!5, inner sep=5pt,
      text=black
  },
  arrow/.style={->, thick, draw=black, -{Stealth[length=3mm]}}
]

% Top row: full model
\node[box, fill=red!70!black!10, text=black] (full) {Relaxation model with\\ \textbf{Standard Linear Solid} viscoelastic constitutive law};

% Second row: constituent models
\node[box, below=1.5cm of full, xshift=-3.5cm, fill=blue!10] (elastic) {Model with\\ \textbf{elastic} constitutive law};
\node[box, below=1.5cm of full, xshift=3.5cm, fill=green!70!black!10] (rigid) {Model with\\ \textbf{Kelvin-Voigt} viscoelastic constitutive law};

% Connections using L-shaped arrows with labels above
%\draw[arrow] (full.west) -| (elastic.north) 
%    node[pos=0.75,left,align=right, text=black]{$\tau_r \to 0$ \\ $\eta \to 0$}
%    node[above, midway, xshift=0.1cm, text=black]{\textit{Hyperbolic scaling}};
\draw[arrow] (full.west) -| (elastic.north) 
    node[pos=0.75,left,align=right, text=black]{$\tau_r \to 0$ \\ while $\eta \to 0$\\ or $E_0 \to E_\infty$}
    node[above, midway, xshift=-0.3cm, text=black]{\textit{Hyperbolic scaling}};
\draw[arrow] (full.east) -| (rigid.north) 
    node[pos=0.75,right,align=left, text=black]{$\tau_r \to 0$ \\while $E_0 \to \infty$}
    node[midway, above, xshift=0.2cm, text=black]{\textit{Diffusive scaling}};

\end{tikzpicture}
%\end{adjustbox}
\caption{Asymptotic limits of the multiscale constitutive framework. By appropriately scaling the viscoelastic parameters, the SLS rheological law allows the blood flow model to capture two regimes: purely elastic behavior with hyperbolic dynamics or Kelvin-Voigt viscoelastic behavior with diffusive dynamics.}
\label{fig:multi}
\end{figure*}
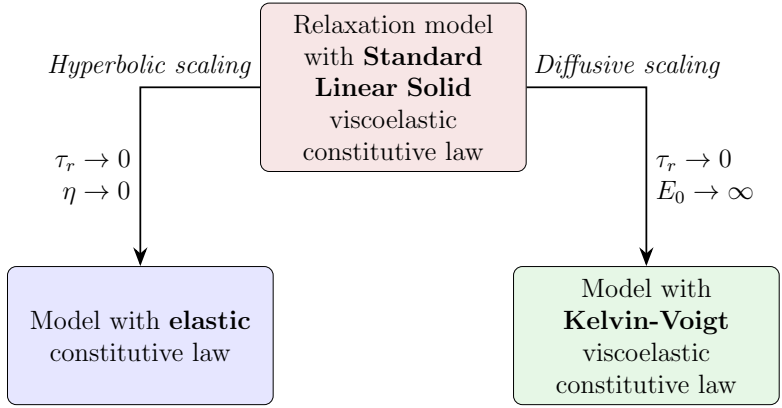

We now examine the asymptotic behavior of system~\eqref{eq:visc_model} in the limit of vanishing relaxation time, $\tau_r \to 0$, commonly referred to as the zero-relaxation limit. In this limit, with suitably chosen scaling parameters, the model can capture different mechanical responses of the vessel laws, as summarized in Figure~\ref{fig:multi} and discussed in the following. For a more detailed description of the multiscale limits the reader can refer to \cite{bertaglia2023multiscale}.

\paragraph{Hyperbolic scaling}\label{sec:2.1.2.1}
In the limit $\tau_r \to 0$ while $\eta \to 0$ or equivalently $E_0 \to E_\infty$, the relaxation function~\eqref{eq:relaxation} reduces to $E(t) \to E_\infty$, so that the vessel wall behaves as a purely elastic material with Young modulus $E_\infty$. Consequently, the viscoelastic tube law reduces to the classical algebraic elastic constitutive law
\begin{equation}
p = F(A) = p_0 + \frac{E_\infty}{W}\left(\alpha^m - \alpha^n\right),
\end{equation}
and the governing equations recover the standard 1D elastic blood flow system:
\begin{align}
&\frac{\partial A}{\partial t} + \frac{\partial (A u)}{\partial x} = 0, \\
&\frac{\partial (A u)}{\partial t} + \frac{\partial (A u^2)}{\partial x} + \frac{A}{\rho}\frac{\partial F(A)}{\partial x} = 0.
\label{eq:hyperbolic}
\end{align}

\paragraph{Diffusive scaling}
In the limit $\tau_r \to 0$ and $E_0 \to \infty$, while keeping $\eta \sim \tau_r E_0$ finite, the relaxation equation~\eqref{eq:visc_model} leads to a diffusive behavior described by the Kelvin-Voigt constitutive law~\cite{lakes2009viscoelastic}:
\begin{equation}
\begin{aligned}
p &= F(A) - \eta G(A) \frac{\partial (A u)}{\partial x} \\[1ex]
&= p_0 + \frac{E_\infty}{W}(\alpha^m - \alpha^n) - \frac{\eta}{W A} (m \alpha^m - n \alpha^n) \frac{\partial (A u)}{\partial x}.
\end{aligned}
\end{equation}
Substituting this relation into the momentum equation~\eqref{eq:visc_model}, the system becomes parabolic:
\begin{align}
&\frac{\partial A}{\partial t} + \frac{\partial (A u)}{\partial x} = 0, \\
&\frac{\partial (A u)}{\partial t} + \frac{\partial (A u^2)}{\partial x} + \frac{A}{\rho}\frac{\partial F(A)}{\partial x} 
= \frac{A}{\rho} \frac{\partial}{\partial x} \Big(\eta G(A) \frac{\partial (A u)}{\partial x}\Big).
\end{align}
The additional parabolic term in the momentum equation represents viscous dissipation of the vessel wall over long time scales. This asymptotic regime exactly recovers the form of 1D blood flow models based on the Kelvin-Voigt rheological law~\cite{lakes2009viscoelastic}.

{\color{black}
\paragraph{Physiological relevance of the asymptotic regimes}
To contextualize these theoretical limits, it is useful to evaluate how the scaling parameters relate to physiological conditions. Because mechanical parameters vary considerably with vessel type, age, and pathophysiological state, a direct correspondence between these mathematical limits and specific physiological conditions is difficult to establish. Nevertheless, their physiological relevance can be assessed by considering under which conditions realistic vascular parameter values approach the corresponding asymptotic limits.
%Nevertheless, the asymptotic regimes provide meaningful approximations of physiological vascular dynamics.

Regarding the relaxation time $\tau_r$, it is typically of the order of $10^{-4}$--$10^{-3}~\mathrm{s}$, i.e., approximately three to four orders of magnitude smaller than the characteristic time scale of blood flow in large vessels ($\sim 1~\mathrm{s}$) \cite{piccioli2021}. Consequently, physiological values of $\tau_r$ are already relatively close to the zero-relaxation limit ($\tau_r \to 0$) shared by both asymptotic regimes.

For the instantaneous Young's modulus $E_0$, it is generally of the order of $10^6~\mathrm{Pa}$ and can reach substantially higher values under pathological conditions associated with increased vessel stiffness, such as hypertension and atherosclerosis~\cite{celant2023modeling}. Thus, the large $E_0$ regime considered in the diffusive scaling is relevant for stiffened vessels, although the limit $E_0 \to \infty$ remains an asymptotic idealization.

Finally, regarding the wall viscosity $\eta$, the hyperbolic limit ($\eta = 0~\mathrm{Pa\cdot~s}$) corresponds to a purely elastic vessel wall. Although setting $\eta = 0~\mathrm{Pa\cdot~s}$ is not strictly physiologically exact, this assumption is widely adopted in the literature to simplify the mathematical description of vascular blood flow while still capturing the main features of blood flow dynamics \cite{alastruey2012arterial}. Thus, although idealized, the $\eta = 0~\mathrm{Pa\cdot~s}$ case represents a commonly used approximation of vascular behavior.
}

\subsection{Asymptotic-Preserving Neural Networks}\label{sec3}

Let us consider a system of PDEs defined over a spatio-temporal domain $\Omega \subset \mathbb{R}^d \times \mathbb{R}$ and depending on a set of physical parameters $\boldsymbol{\xi}$. In compact form, the governing equations and the associated initial and boundary conditions can be written as
\begin{align}
&\mathcal{F}(\mathbf{U}, \mathbf{x}, t; \boldsymbol{\xi}) = 0, \qquad (\mathbf{x}, t) \in \Omega, \\
&\mathcal{B}(\mathbf{U}, \mathbf{x}, t; \boldsymbol{\xi}) = 0, \qquad (\mathbf{x}, t) \in \partial \Omega,
\label{eq:residuals}
\end{align}
where $\mathcal{F}$ denotes the differential operator associated with the PDEs system and $\mathcal{B}$ represents the operator enforcing the conditions on the domain boundary $\partial \Omega$, which encompasses both spatial boundaries and the initial time $t=0$.  The function $\mathbf{U}(\mathbf{x},t)$ denotes the vector of unknown solution variables over $\Omega$.

In many practical applications, the governing PDEs depend on physical parameters $\boldsymbol{\xi}$ that are difficult to measure or estimate accurately. This is particularly relevant for the viscoelastic model~\eqref{eq:visc_model}, where the viscoelastic coefficients are difficult to identify from experimental measurements and are associated with high uncertainties. As a result, classical numerical integration methods can be limited by parameter uncertainty, motivating the use of data-driven surrogate models that learn a direct mapping from input variables to the corresponding solution fields of the underlying PDEs.

Among these approaches, Neural Networks (NNs) have shown strong performance due to their ability to approximate complex nonlinear mappings~\cite{goodfellow2016deep, cabini2024fast}. In this setting, the solution $\mathbf{U}(\mathbf{x},t)$ is approximated by a NN that takes the space--time coordinates $(\mathbf{x},t)$ as input and outputs the predicted state variables:
\begin{equation}
\hat{\mathbf{U}}(\mathbf{x},t;\boldsymbol{\theta}) \approx \mathbf{U}(\mathbf{x},t),
\end{equation}
where $\boldsymbol{\theta}$ denotes the set of trainable parameters.

This Section provides an overview of the main characteristics of NNs~\cite{goodfellow2016deep}, PINNs~\cite{raissi2019physics, karniadakis2021physics}, and their extension to APNNs~\cite{jin2023asymptotic, bertaglia2021asymptotic, bertaglia2022asymptotic} in the context of PDEs solution approximation.

\subsubsection{Neural Networks (NNs)}

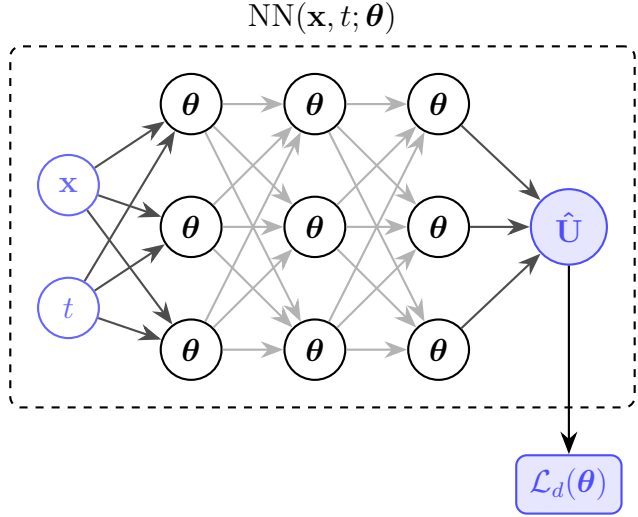
\begin{figure*}
\centering
%\begin{adjustbox}{width=0.55\textwidth}
    \begin{tikzpicture}[
    node distance=1.4cm,
    thick,
    neuron/.style={circle,draw=black,minimum size=8mm,text=black},
    input/.style={circle,draw,blue!60,minimum size=8mm},
    output/.style={circle,draw,blue!70,fill=blue!10,minimum size=10mm},
    box/.style={rectangle,draw=black,dashed,rounded corners,inner sep=10pt},
    arrow/.style={-{Stealth[length=3mm]}}
]

% FIRST HIDDEN LAYER ------------------------------------------------
\node[neuron] (h11) {$\boldsymbol{\theta}$};
\node[neuron,below=0.8cm of h11] (h12) {$\boldsymbol{\theta}$};
\node[neuron,below=0.8cm of h12] (h13) {$\boldsymbol{\theta}$};

% INPUT LAYER — centered vertically w.r.t h11–h13
\node[input, left=1.2cm of $(h11)!0.33!(h13)$] (x) {$\mathbf{x}$};
\node[input, below=0.8cm of x] (t) {$t$};

% SECOND HIDDEN LAYER ----------------------------------------------
\node[neuron,right=0.8cm of h11] (h21) {$\boldsymbol{\theta}$};
\node[neuron,below=0.8cm of h21] (h22) {$\boldsymbol{\theta}$};
\node[neuron,below=0.8cm of h22] (h23) {$\boldsymbol{\theta}$};

% THIRD HIDDEN LAYER ------------------------------------------------
\node[neuron,right=0.8cm of h21] (h31) {$\boldsymbol{\theta}$};
\node[neuron,below=0.8cm of h31] (h32) {$\boldsymbol{\theta}$};
\node[neuron,below=0.8cm of h32] (h33) {$\boldsymbol{\theta}$};

% OUTPUT — centered vertically w.r.t h31–h33
\node[output, right=1.2cm of $(h31)!0.5!(h33)$] (u) {$\mathbf{\hat{U}}$};

% CONNECTIONS INPUT → H1
\foreach \a in {h11,h12,h13}
{
    \draw[arrow,black!70] (x) -- (\a);
    \draw[arrow,black!70] (t) -- (\a);
}

% CONNECTIONS H1 → H2
\foreach \a in {h11,h12,h13}
\foreach \b in {h21,h22,h23}
    \draw[arrow,gray!60] (\a) -- (\b);

% CONNECTIONS H2 → H3
\foreach \a in {h21,h22,h23}
\foreach \b in {h31,h32,h33}
    \draw[arrow,gray!60] (\a) -- (\b);

% CONNECTIONS H3 → OUTPUT
\foreach \a in {h31,h32,h33}
    \draw[arrow,black!70] (\a) -- (u);

% BOX AROUND NN ----------------------------------------------------
\node[box,fit=(x)(t)(h11)(h13)(h31)(h33)(u),label={[text=black,yshift=0.cm]NN$(\mathbf{x},t;\boldsymbol{\theta})$}] (NNbox) {};

\node[rectangle, rounded corners, draw, below=2.5cm of u, inner sep=5pt, blue!70,fill=blue!10] (Ld) {$\mathcal{L}_d(\boldsymbol{\theta})$};

\draw[arrow] (u) -- (Ld);
\end{tikzpicture}
%\end{adjustbox}
\caption{Feedforward Neural Network architecture. The network maps space--time inputs $(\mathbf{x},t)$ to the predicted solution $\mathbf{U}$, with parameters $\boldsymbol{\theta}$ optimized by minimizing the loss $\mathcal{L}_d(\boldsymbol{\theta})$.}
\label{fig:NN}
\end{figure*}

A fully-connected feedforward NN, which forms the basis of PINNs and APNNs, consists of $L$ layers and computes the input--output mapping recursively as
\begin{equation}
\mathbf{h}^{(0)} = (\mathbf{x},t), \quad
\mathbf{h}^{(l)} = \sigma\left( \mathbf{W}^{(l)} \mathbf{h}^{(l-1)} + \mathbf{b}^{(l)} \right), \quad l = 1, \dots, L,
\end{equation}
where $\mathbf{W}^{(l)}$ and $\mathbf{b}^{(l)}$ are the weight matrices and bias vectors, and $\sigma(\cdot)$ is a component-wise activation function. The output of the final layer defines the network approximation, $\hat{\mathbf{U}} = \mathbf{h}^{(L)}$ (Figure~\ref{fig:NN}).

The network parameters $\boldsymbol{\theta}$ are determined by minimizing a loss function $\mathcal{L}_d$ that quantifies the discrepancy between predictions and available reference data. A common choice is the Mean Squared Error:
\begin{equation}
\mathcal{L}_d(\boldsymbol{\theta}) = \frac{1}{N_d} \sum_{n=1}^{N_d} \left\| \hat{\mathbf{U}}(\mathbf{x}_n, t_n; \boldsymbol{\theta}) - \mathbf{U}_n \right\|^2,
\end{equation}
where $\{(\mathbf{x}_n,t_n,\mathbf{U}_n)\}_{n=1}^{N_d}$ is the training set, assumed to adequately cover the spatio-temporal domain $\Omega$. The optimal parameters are then given by:
\begin{equation}
\boldsymbol{\theta}^\ast = \mathop{\mathrm{argmin}}\limits_{\boldsymbol{\theta}} \mathcal{L}_d(\boldsymbol{\theta}).
\end{equation}
Although this approach can achieve good performance in approximating the PDEs solution $\mathbf{U}(\mathbf{x},t)$, it has three main limitations~\cite{raissi2019physics, karniadakis2021physics}. First, in many real-world applications, obtaining a dense and representative training set that adequately covers $\Omega$ is difficult, costly, or impractical. Second, predictions at locations distant from the training data may be unreliable, potentially causing errors in unsampled regions. Third, purely data-driven networks do not embed physical knowledge, so their predictions can violate fundamental laws and produce inconsistent solutions.

\subsubsection{Physics-Informed Neural Networks (PINNs)}

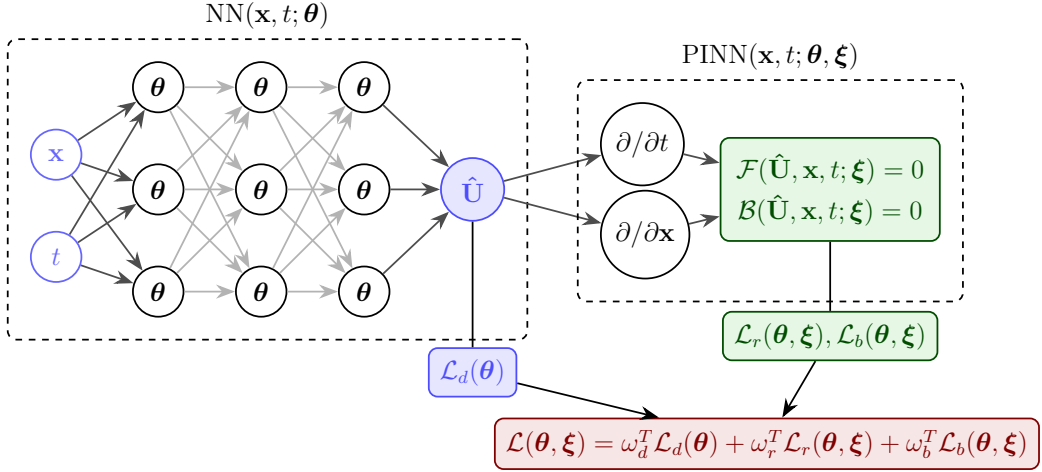
\begin{figure*}
\centering
%\begin{adjustbox}{width=\textwidth}
    \begin{tikzpicture}[
    node distance=1.4cm,
    thick,
    neuron/.style={circle,draw=black,minimum size=8mm,text=black},
    input/.style={circle,draw,blue!60,minimum size=8mm},
    output/.style={circle,draw,blue!70,fill=blue!10,minimum size=10mm},
    box/.style={rectangle,draw=black,dashed,rounded corners,inner sep=10pt},
    arrow/.style={-{Stealth[length=3mm]}}
]

% FIRST HIDDEN LAYER ------------------------------------------------
\node[neuron] (h11) {$\boldsymbol{\theta}$};
\node[neuron,below=0.8cm of h11] (h12) {$\boldsymbol{\theta}$};
\node[neuron,below=0.8cm of h12] (h13) {$\boldsymbol{\theta}$};

% INPUT LAYER — centered vertically w.r.t h11–h13
\node[input, left=1.2cm of $(h11)!0.33!(h13)$] (x) {$\mathbf{x}$};
\node[input, below=0.8cm of x] (t) {$t$};

% SECOND HIDDEN LAYER ----------------------------------------------
\node[neuron,right=0.8cm of h11] (h21) {$\boldsymbol{\theta}$};
\node[neuron,below=0.8cm of h21] (h22) {$\boldsymbol{\theta}$};
\node[neuron,below=0.8cm of h22] (h23) {$\boldsymbol{\theta}$};

% THIRD HIDDEN LAYER ------------------------------------------------
\node[neuron,right=0.8cm of h21] (h31) {$\boldsymbol{\theta}$};
\node[neuron,below=0.8cm of h31] (h32) {$\boldsymbol{\theta}$};
\node[neuron,below=0.8cm of h32] (h33) {$\boldsymbol{\theta}$};

% OUTPUT — centered vertically w.r.t h31–h33
\node[output, right=1.2cm of $(h31)!0.5!(h33)$] (u) {$\mathbf{\hat{U}}$};

% CONNECTIONS INPUT → H1
\foreach \a in {h11,h12,h13}
{
    \draw[arrow,black!70] (x) -- (\a);
    \draw[arrow,black!70] (t) -- (\a);
}

% CONNECTIONS H1 → H2
\foreach \a in {h11,h12,h13}
\foreach \b in {h21,h22,h23}
    \draw[arrow,gray!60] (\a) -- (\b);

% CONNECTIONS H2 → H3
\foreach \a in {h21,h22,h23}
\foreach \b in {h31,h32,h33}
    \draw[arrow,gray!60] (\a) -- (\b);

% CONNECTIONS H3 → OUTPUT
\foreach \a in {h31,h32,h33}
    \draw[arrow,black!70] (\a) -- (u);

% BOX AROUND NN ----------------------------------------------------
\node[box,fit=(x)(t)(h11)(h13)(h31)(h33)(u),label={[text=black,yshift=0.cm]NN$(\mathbf{x},t;\boldsymbol{\theta})$}] (NNbox) {};

% AUTODIFF BLOCK ---------------------------------------------------
% distanza verticale tra dt e dx
\def\gap{1.2cm}

% AUTODIFF BLOCK centrato rispetto a u
\node[draw=black, text=black, circle, minimum size=10mm, right=1.5cm of u, yshift=0.6*\gap] (dt) {$\partial/\partial t$};
\node[draw=black, text=black, circle, minimum size=10mm, right=1.5cm of u, yshift=-0.6*\gap] (dx) {$\partial/\partial \mathbf{x}$};

\draw[arrow,black!70] (u) -- (dt);
\draw[arrow,black!70] (u) -- (dx);

% PDE BOX ----------------------------------------------------------
\node[rectangle, draw, rounded corners, right=1.2cm of $(dx)!0.5!(dt)$, inner sep=7pt, align=center, green!30!black!100, fill=green!70!black!10] (PDE) 
{
  $\mathcal{F}(\mathbf{\hat{U}},\mathbf{x},t;\boldsymbol{\xi})=0$\\[4pt]
  $\mathcal{B}(\mathbf{\hat{U}},\mathbf{x},t;\boldsymbol{\xi})=0$
};

\draw[arrow,black!70] (dt) -- (PDE);
\draw[arrow,black!70] (dx) -- (PDE);

% LOSS FUNCTION -----------------------------------------------------
\node[rectangle, rounded corners, draw, thick, below=2.8cm of PDE, xshift=-1cm, inner sep=5pt, red!50!black!100, fill=red!70!black!10] (loss)
{$\mathcal{L}(\boldsymbol{\theta},\boldsymbol{\xi})=\omega_d^T\mathcal{L}_d(\boldsymbol{\theta}) 
+ \omega_r^T\mathcal{L}_r(\boldsymbol{\theta},\boldsymbol{\xi}) 
+ \omega_b^T\mathcal{L}_b(\boldsymbol{\theta},\boldsymbol{\xi})$};

\node[rectangle, rounded corners, draw, below=2.cm of u, inner sep=5pt, blue!70,fill=blue!10] (Ld) {$\mathcal{L}_d(\boldsymbol{\theta})$};
\node[rectangle, rounded corners, draw, below=1.1cm of PDE, inner sep=5pt, green!30!black!100, fill=green!70!black!10] (Lrb) {$\mathcal{L}_r(\boldsymbol{\theta},\boldsymbol{\xi}), \mathcal{L}_b(\boldsymbol{\theta},\boldsymbol{\xi})$};

\draw[arrow] (u) -- (Ld) -- (loss);
\draw[arrow] (PDE) -- (Lrb) -- (loss);

% BOX AROUND PINN --------------------------------------------------
\node[box,fit=(dt)(dx)(PDE),label={[text=black,yshift=0.cm]PINN$(\mathbf{x},t;\boldsymbol{\theta},\boldsymbol{\xi})$}] (PINNbox) {};

% TRAINING LOOP -----------------------------------------------------
%\node[diamond, draw, below=1cm of loss, aspect=2, inner sep=2pt] (check)
%{$<\varepsilon_{\rm err}?$};

%\draw[arrow] (loss) -- (check);

%\node[right=1.4cm of check, red] (no) {no};
%\draw[arrow] (check) -- (no);

%\node[left=1.4cm of check, green!50!black] (yes) {yes};
%\draw[arrow] (check) -- (yes);

%\draw[arrow] (no) |- ([yshift=-5cm]t.south) -- (t.south);

%\node[below=0.7cm of yes] {End};

\end{tikzpicture}
%\end{adjustbox}
\caption{PINN architecture. The network incorporates both data-driven and physics-based constraints in the loss function, including the data mismatch term $\mathcal{L}_d$, the PDEs residual term $\mathcal{L}_r$, and the boundary or initial condition term $\mathcal{L}_b$, to ensure consistency with the governing equations.}
\label{fig:PINN}
\end{figure*}

PINNs overcome the above-mentioned limitations by embedding physical knowledge into the learning process, ensuring that NN predictions satisfy the governing physical laws~\cite{raissi2019physics, karniadakis2021physics}.

To transform a standard NN into a PINN (Figure~\ref{fig:PINN}), the loss function is augmented to include both data and physics constraints~\eqref{eq:residuals}:
\begin{equation}
\mathcal{L}(\boldsymbol{\theta}, \boldsymbol{\xi}) = \omega_d \mathcal{L}_d(\boldsymbol{\theta}, \boldsymbol{\xi}) + \omega_r \mathcal{L}_r(\boldsymbol{\theta}, \boldsymbol{\xi}) + \omega_b \mathcal{L}_b(\boldsymbol{\theta}, \boldsymbol{\xi}),
\label{eq:loss_pinn}
\end{equation}
where $\mathcal{L}_r$ and $\mathcal{L}_b$ measure the residuals of the differential operator and initial or boundary conditions at scattered collocations points, $\{(\mathbf{x}_n^r,t_n^r)\}_{n=1}^{N_r}$ and $\{(\mathbf{x}_n^b,t_n^b)\}_{n=1}^{N_b}$, respectively in $\Omega$ and $\partial \Omega$:
\begin{align}
\mathcal{L}_r(\boldsymbol{\theta}, \boldsymbol{\xi}) &= \frac{1}{N_r} \sum_{n=1}^{N_r} \big\| \mathcal{F}(\hat{\mathbf{U}}(\mathbf{x}_n^r, t_n^r; \boldsymbol{\theta}), \mathbf{x}_n^r, t_n^r; \boldsymbol{\xi}) \big\|^2, \\
\mathcal{L}_b(\boldsymbol{\theta}, \boldsymbol{\xi}) &= \frac{1}{N_b} \sum_{n=1}^{N_b} \big\| \mathcal{B}(\hat{\mathbf{U}}(\mathbf{x}_n^b, t_n^b; \boldsymbol{\theta}), \mathbf{x}_n^b, t_n^b; \boldsymbol{\xi}) \big\|^2.
\end{align}
Gradients required for these loss terms are computed via automatic differentiation, which allows exact computation of derivatives of the NN output with respect to its inputs. The weights $\omega_d$, $\omega_r$, and $\omega_b$ control the relative importance of the data, PDEs residual, and initial or boundary residual terms, respectively.

In forward problems, where $\boldsymbol{\xi}$ is known, the network is trained by
\begin{equation}
\boldsymbol{\theta}^\ast = \mathop{\mathrm{argmin}}\limits_{\boldsymbol{\theta}} \mathcal{L}(\boldsymbol{\theta}, \boldsymbol{\xi}).
\end{equation}
In inverse problems, where $\boldsymbol{\xi}$ is unknown, it is treated as a learnable parameter alongside $\boldsymbol{\theta}$:
\begin{equation}
(\boldsymbol{\theta}^\ast, \boldsymbol{\xi}^\ast) = \mathop{\mathrm{argmin}}\limits_{\boldsymbol{\theta}, \boldsymbol{\xi}} \mathcal{L}(\boldsymbol{\theta}, \boldsymbol{\xi}).
\end{equation}
This allows the PINN to simultaneously infer the physical parameters and approximate the solution.

This training strategy allows PINNs not only to satisfy the governing physical laws but also to generalize more effectively than vanilla NNs, potentially providing accurate predictions even in regions with sparse or no training data~\cite{karniadakis2021physics}.

\subsubsection{Asymptotic-Preserving Neural Networks (APNNs)}
In multiscale problems, such as model \eqref{eq:visc_model}, the governing equations depend on scaling parameters that characterize different physical regimes. In such cases, the learning framework should remain accurate both for the full order model and also for the corresponding asymptotic reduced order models.

APNNs extend the PINN framework to meet this requirement~\cite{jin2023asymptotic, bertaglia2021asymptotic, bertaglia2022asymptotic}. Inspired by AP numerical schemes \cite{jin2022}, APNNs are designed so that the physics-based loss remains consistent with the asymptotic limit of the governing equations, by appropriately reformulating the residuals to respect these limits.

\begin{figure*}
\centering
\begin{tikzpicture}[
    font=\normalsize,
    % distanza verticale e orizzontale tra nodi
    node distance=2.2cm and 4.2cm, % Aumentato leggermente lo spazio orizzontale per far respirare le scritte più grandi
    box/.style={rectangle, rounded corners, minimum width=0.6cm,
                minimum height=0.6cm, align=center, text=black},%, fill=blue!30},
    arr/.style={->, thick, -{Stealth[length=3mm]},
                shorten <=4pt, shorten >=4pt, draw=black}, % spazio uniforme dai box
    lab/.style={fill=white, inner sep=2pt} % <--- Tolto font=\small per seguire la dimensione globale
]

% --- NODES --------------------------------------------------------
\node[box] (RFeps) {$\mathcal{L}_r^{\varepsilon}$};
\node[box, right=of RFeps] (RF0) {$\mathcal{L}_r^{0}$};

\node[box, below=of RFeps] (Feps) {$\mathcal{F}^{\varepsilon}$};
\node[box, below=of RF0]   (F0)   {$\mathcal{F}^{0}$};

% --- HORIZONTAL ARROWS -------------------------------------------
\draw[arr] (RFeps) -- node[text=black, above=8pt, lab] {$\varepsilon \to 0$} (RF0);
\draw[arr] (Feps) -- node[text=black, above=8pt, lab] {$\varepsilon \to 0$} (F0);

% --- VERTICAL ARROWS ---------------------------------------------
\draw[arr] (RFeps) -- node[text=black, right=6pt, lab] {$\mathcal{L}_r \to 0$} (Feps);
\draw[arr] (RF0) -- node[text=black, right=6pt, lab] {$\mathcal{L}_r \to 0$} (F0);

\end{tikzpicture}
\caption{Schematic illustration of the concept of AP property in an APNN. The PINN approximates the solution of the multiscale problem $\mathcal{F}^\varepsilon$ by minimizing the residual $\mathcal{L}_r^\varepsilon$. In the asymptotic limit $\varepsilon \to 0$, the residual converges to $\mathcal{L}_r^0$, which is consistent with the reduced system $\mathcal{F}^0$, ensuring the network preserves the correct asymptotic behavior.}
\label{fig:APNN}
\end{figure*}
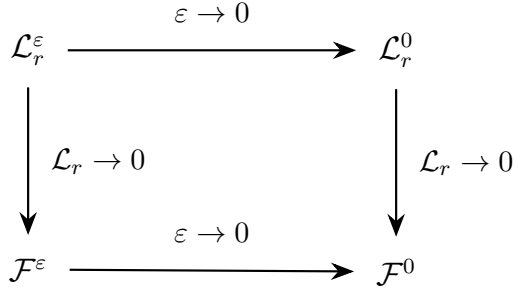

Specifically, consider a multiscale system parametrized by $\varepsilon$, with dynamics described by a differential operator $\mathcal{F}^\varepsilon$. Its solution is approximated by a PINN trained to minimize the physics-based residual $\mathcal{L}_r^\varepsilon$.  
The network is called an APNN if, in the limit $\varepsilon \to 0$, the residual converges to that of the reduced model:
\begin{equation}
\mathcal{L}_r^\varepsilon(\boldsymbol{\theta}) \to \mathcal{L}_r^0(\boldsymbol{\theta}) \quad \text{if } \varepsilon \to 0
\end{equation}
where $\mathcal{L}_r^0$ is the residual corresponding to the asymptotic operator $\mathcal{F}^0$ (see Figure~\ref{fig:APNN}).  

\subsection{APNN applied to blood flow modeling}\label{sec4}

\begin{figure*}
\centering
\begin{adjustbox}{width=\textwidth}
    \begin{tikzpicture}[
    node distance=1.4cm,
    thick,
    neuron/.style={circle,draw=black,minimum size=8mm,text=black},
    input/.style={circle,draw,blue!60,minimum size=8mm},
    output/.style={circle,draw,blue!70,fill=blue!10,minimum size=10mm},
    box/.style={rectangle,draw=black,dashed,rounded corners,inner sep=10pt},
    arrow/.style={-{Stealth[length=3mm]}}
]

% FIRST HIDDEN LAYER ------------------------------------------------
\node[neuron] (h11) {$\boldsymbol{\theta}$};
\node[neuron,below=0.8cm of h11] (h12) {$\boldsymbol{\theta}$};
\node[neuron,below=0.8cm of h12] (h13) {$\boldsymbol{\theta}$};

% INPUT LAYER — centered vertically w.r.t h11–h13
\node[input, left=1.2cm of $(h11)!0.33!(h13)$] (x) {$x$};
\node[input, below=0.8cm of x] (t) {$t$};

% SECOND HIDDEN LAYER ----------------------------------------------
\node[neuron,right=0.8cm of h11] (h21) {$\boldsymbol{\theta}$};
\node[neuron,below=0.8cm of h21] (h22) {$\boldsymbol{\theta}$};
\node[neuron,below=0.8cm of h22] (h23) {$\boldsymbol{\theta}$};

% THIRD HIDDEN LAYER ------------------------------------------------
\node[neuron,right=0.8cm of h21] (h31) {$\boldsymbol{\theta}$};
\node[neuron,below=0.8cm of h31] (h32) {$\boldsymbol{\theta}$};
\node[neuron,below=0.8cm of h32] (h33) {$\boldsymbol{\theta}$};

% OUTPUT — centered vertically w.r.t h31–h33
\node[output,right=1.2cm of h31] (a) {$\hat{A}$};
\node[output,below=0.8cm of a] (u) {$\hat{u}$};
\node[output,below=0.8cm of u] (p) {$\hat{p}$};

% CONNECTIONS INPUT → H1
\foreach \a in {h11,h12,h13}
{
    \draw[arrow,black!70] (x) -- (\a);
    \draw[arrow,black!70] (t) -- (\a);
}

% CONNECTIONS H1 → H2
\foreach \a in {h11,h12,h13}
\foreach \b in {h21,h22,h23}
    \draw[arrow,gray!60] (\a) -- (\b);

% CONNECTIONS H2 → H3
\foreach \a in {h21,h22,h23}
\foreach \b in {h31,h32,h33}
    \draw[arrow,gray!60] (\a) -- (\b);

% CONNECTIONS H3 → OUTPUT
\foreach \a in {h31,h32,h33}
\foreach \b in {a,u,p}
    \draw[arrow,black!70] (\a) -- (\b);

% BOX AROUND OUTPUT ----------------------------------------------------
\begin{scope}[on background layer]
\node[rectangle, fit=(a)(u)(p), rounded corners, draw=blue!70, fill=blue!10, inner sep=6pt] (OUTbox) {};
% Sovrapposizione su P
\node[rectangle, fill=white, rounded corners, draw=blue!70, dotted, fit=(p)] {};
\end{scope}

% BOX AROUND NN ----------------------------------------------------
\node[box,fit=(x)(t)(h11)(h13)(h31)(h33)(a)(u)(p)(OUTbox),label={[yshift=0.cm,text=black]NN$(x,t;\boldsymbol{\theta})$}] (NNbox) {};

% AUTODIFF BLOCK ---------------------------------------------------
% distanza verticale tra dt e dx
\def\gap{1.2cm}

% AUTODIFF BLOCK centrato rispetto a u
\node[draw=black, circle, minimum size=10mm, right=1.5cm of OUTbox, yshift=0.6*\gap,text=black] (dt) {$\partial/\partial t$};
\node[draw=black, circle, minimum size=10mm, right=1.5cm of OUTbox, yshift=-0.6*\gap,text=black] (dx) {$\partial/\partial x$};

\draw[arrow,black!70] (OUTbox) -- (dt);
\draw[arrow,black!70] (OUTbox) -- (dx);

% PDE BOX ----------------------------------------------------------
\node[rectangle, draw, rounded corners, right=1.2cm of $(dx)!0.5!(dt)$, inner sep=7pt, align=center, green!30!black!100, fill=green!70!black!10] (PDE) 
{
\(\begin{aligned}
    & \frac{\partial \hat{A}}{\partial t} + \frac{\partial (\hat{A}\hat{u})}{\partial x} = 0, \\
    & \frac{\partial (\hat{A}\hat{u})}{\partial t} + \frac{\partial (\hat{A}\hat{u}^2)}{\partial x} + \frac{\hat{A}}{\rho}  \frac{\partial \hat{p}}{\partial x} = 0, \\
    & \tau_r\left(\frac{\partial \hat{p}}{\partial t} + E_0 G(\hat{A}) \frac{\partial (\hat{A}\hat{u})}{\partial x}\right) +(\hat{p}-F(\hat{A}))=0\\[15pt]
    & \mathcal{B}(\hat{A},\hat{u},\hat{p},x,t;\tau_r, E_0)=0
\end{aligned}\)
};

\draw[arrow,black!70] (dt) -- (PDE);
\draw[arrow,black!70] (dx) -- (PDE);

% LOSS FUNCTION -----------------------------------------------------
\node[rectangle, rounded corners, draw, thick, below=2.8cm of PDE, xshift=-1cm, inner sep=5pt, red!50!black!100, fill=red!70!black!10] (loss)
{$\mathcal{L}(\boldsymbol{\theta},\tau_r, E_0)=\omega_d^T\mathcal{L}_d(\boldsymbol{\theta}) 
+ \omega_r^T\mathcal{L}_r(\boldsymbol{\theta},\tau_r, E_0) 
+ \omega_b^T\mathcal{L}_b(\boldsymbol{\theta},\tau_r, E_0)$};

\node[rectangle, rounded corners, draw, below=1.cm of OUTbox, inner sep=5pt, blue!70,fill=blue!10] (Ld) {$\mathcal{L}_d(\boldsymbol{\theta})$};
\node[rectangle, rounded corners, draw, below=1.1cm of PDE, inner sep=5pt, green!30!black!100, fill=green!70!black!10] (Lrb) {$\mathcal{L}_r(\boldsymbol{\theta},\tau_r, E_0), \mathcal{L}_b(\boldsymbol{\theta},\tau_r, E_0)$};

\draw[arrow] (OUTbox) -- (Ld) -- (loss);
\draw[arrow] (PDE) -- (Lrb) -- (loss);

% BOX AROUND PINN --------------------------------------------------
\node[box,fit=(dt)(dx)(PDE),label={[yshift=0.cm,text=black]APNN$(x,t;\boldsymbol{\theta},\tau_r, E_0$)}] (PINNbox) {};

% TRAINING LOOP -----------------------------------------------------
%\node[diamond, draw, below=1cm of loss, aspect=2, inner sep=2pt] (check)
%{$<\varepsilon_{\rm err}?$};

%\draw[arrow] (loss) -- (check);

%\node[right=1.4cm of check, red] (no) {no};
%\draw[arrow] (check) -- (no);

%\node[left=1.4cm of check, green!50!black] (yes) {yes};
%\draw[arrow] (check) -- (yes);

%\draw[arrow] (no) |- ([yshift=-5cm]t.south) -- (t.south);

%\node[below=0.7cm of yes] {End};

\end{tikzpicture}
\end{adjustbox}
\caption{Schematic representation of the APNN framework. The neural network maps the space--time inputs $(x,t)$ to the predicted fields $(\hat{A}, \hat{u}, \hat{p})$. The data loss $\mathcal{L}_d$ is evaluated only on the observed variables $\hat{A}$ and $\hat{u}$ (highlighted by the blue box), while the pressure $\hat{p}$ is inferred implicitly through the physics-based residual $\mathcal{L}_r$ and boundary and initial losses $\mathcal{L}_b$, which are computed from the blood flow model~\eqref{eq:visc_model}.}
\label{fig:bf_pinn}
\end{figure*}
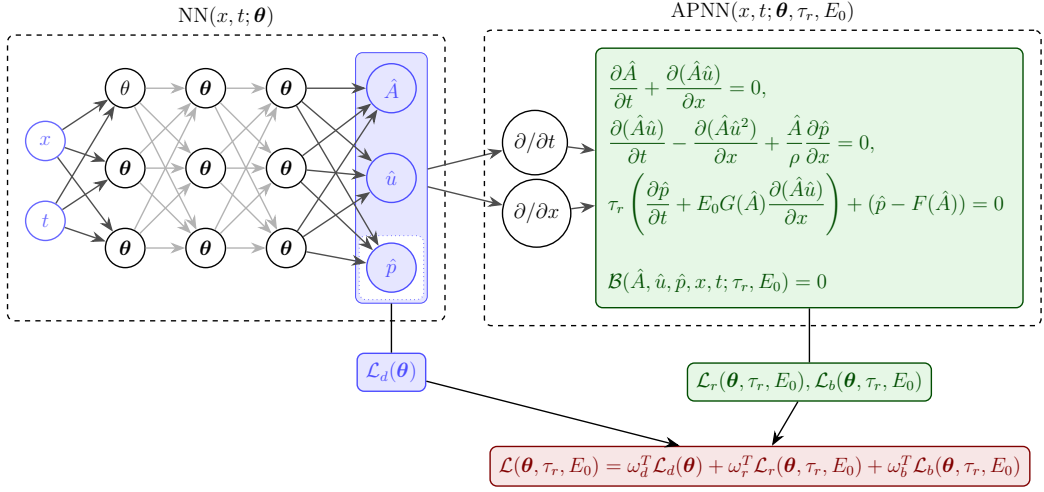

The APNN framework is specialized here to the viscoelastic blood flow model described in Section~\ref{sec2}, with the objective of simultaneously reconstructing the hemodynamic variables and inferring the unknown viscoelastic parameters. A schematic overview of the proposed approach is shown in Figure~\ref{fig:bf_pinn}.

The APNN takes the spatio-temporal coordinates $(x,t)$ as input and returns the predicted state variables
\begin{equation}
\hat{\mathbf{U}}(x,t;\boldsymbol{\theta}) = \big(\hat{A}(x,t;\boldsymbol{\theta}), \hat{u}(x,t;\boldsymbol{\theta}), \hat{p}(x,t;\boldsymbol{\theta})\big),
\end{equation}
which approximate the solution of system~\eqref{eq:visc_model}. In addition to the network parameters $\boldsymbol{\theta}$, the viscoelastic coefficients $\boldsymbol{\xi} = (\tau_r, E_0)$ are treated as trainable variables and optimized jointly during the training process.

The total training loss is defined as in~\eqref{eq:loss_pinn} and consists of a supervised data term and physics-based terms. 
The supervised loss is computed using available measurements of the cross-sectional area and mean axial velocity and is defined as
\begin{equation}
\mathcal{L}_d(\boldsymbol{\theta}) =
\frac{1}{N_d} \sum_{n=1}^{N_d}
\Big(
\big\|\hat{A}(x^d_n,t^d_n;\boldsymbol{\theta})-A_n\big\|^2
+
\big\|\hat{u}(x^d_n,t^d_n;\boldsymbol{\theta})-u_n\big\|^2
\Big),
\label{eq:loss_d}
\end{equation}
where $\{(x^d_n,t^d_n)\}_{n=1}^{N_d}$ denotes the set of measurement locations and $A_n$ and $u_n$ are the corresponding observed values. Notice that the internal pressure is not included in the supervised loss, as pressure measurements are typically available only for superficial vessels and are generally inaccessible for the remaining vessels in vivo. Consequently, the pressure is inferred implicitly during training by enforcing that the pressure predicted by the APNN satisfies the governing physical laws.

The physics-based loss is defined by enforcing the governing equations~\eqref{eq:visc_model} at a set of collocation points $\{(x^r_n,t^r_n)\}_{n=1}^{N_r}$ in the spatio-temporal domain. Specifically, the residual loss is defined as
\begin{equation}
\begin{aligned}
    \mathcal{L}_{r} (\boldsymbol{\theta},\tau_r, E_0) &= \frac{1}{N_r} \sum_{n=1}^{N_r} \big\| \mathcal{R}_1(x_n^r, t_n^r; \tau_r, E_0; \boldsymbol{\theta})\big\|^2 \\
    &+ \frac{1}{N_r} \sum_{n=1}^{N_r} \big\|\mathcal{R}_2(x_n^r, t_n^r; \tau_r, E_0; \boldsymbol{\theta})\big\|^2 \\
    &+ \frac{1}{N_r} \sum_{n=1}^{N_r} \big\|\mathcal{R}_3(x_n^r, t_n^r; \tau_r, E_0; \boldsymbol{\theta})\big\|^2,
\label{eq:loss_r}
\end{aligned}
\end{equation}
where the residual operators correspond to the conservation of mass, momentum, and the viscoelastic constitutive relation in~\eqref{eq:visc_model}, respectively:
\begin{equation}
\begin{aligned}
    \mathcal{R}_1 &= \frac{\partial \hat{A}}{\partial t} + \frac{\partial (\hat{A} \hat{u})}{\partial x}, \\
    \mathcal{R}_2 &= \frac{\partial (\hat{A} \hat{u})}{\partial t} + \frac{\partial (\hat{A} \hat{u}^2)}{\partial x} + \frac{\hat{A}}{\rho}\,\frac{\partial \hat{p}}{\partial x}, \\
    \mathcal{R}_3 &= {\tau_r}\biggl(\frac{\partial \hat{p}}{\partial t} + E_0 G(\hat{A})\,\frac{\partial (\hat{A} \hat{u})}{\partial x}\biggr) + \bigl(\hat{p} - F(\hat{A})\bigr).
\label{eq:apnn_res}
\end{aligned}
\end{equation}
\textcolor{black}{Since $F(\hat{A})$ depends on the equilibrium pressure $p_0$, this parameter must be known to evaluate $\mathcal{R}_3$. In all numerical tests, $p_0$ is therefore treated as a known input (coinciding with the diastolic pressure) and kept fixed throughout the training.}

Positivity constraints and initial conditions are enforced through additional penalty terms in the loss function, while no explicit spatial boundary conditions are imposed. In particular, the positivity of the internal pressure is imposed at selected collocation points $\{(x^p_n,t^p_n)\}_{n=1}^{N_p}$ within the spatio-temporal domain, while the initial conditions are enforced at $t=0$, in $\{(x^i_n,0)\}_{n=1}^{N_i}$:
\begin{equation}
\begin{aligned}
    \mathcal{L}_{b} (\boldsymbol{\theta},\tau_r, E_0) &= %\frac{1}{N_p} \sum_{n=1}^{N_p} \big\| \mathrm{abs}(\hat{A}(x_p^n, t_p^n; \boldsymbol{\theta})) - \hat{A}(x_p^n, t_p^n; \boldsymbol{\theta})\big\|^2 \\
    \frac{1}{N_p} \sum_{n=1}^{N_p} \big\| \mathrm{abs}(\hat{p}(x^p_n, t^p_n; \boldsymbol{\theta})) - \hat{p}(x^p_n, t^p_n; \boldsymbol{\theta})\big\|^2 \\
    &+ \frac{1}{N_i} \sum_{n=1}^{N_i} \big\|\hat{A}(x^i_n,0; \boldsymbol{\theta})-A_0(x^i_n)\big\|^2 \\
    &+ \frac{1}{N_i} \sum_{n=1}^{N_i} \big\|\hat{p}(x^i_n,0; \boldsymbol{\theta})-p_0(x^i_n)\big\|^2.
\label{eq:loss_b}
\end{aligned}
\end{equation}
Notice that no positivity constraint is imposed on the cross-sectional area because the network output for $\hat{A}$ is designed to be strictly positive, which is essential to ensure that the PDE residuals are well-defined and can be computed numerically, as detailed in the next section.

The AP property is enforced by defining the residual term $\mathcal{R}_3$ from the viscoelastic constitutive law in~\eqref{eq:visc_model} and explicitly multiplying it by the relaxation parameter $\tau_r$. This formulation enables the NN to consistently recover the correct asymptotic limits as $\tau_r \to 0$, yielding either the purely elastic hyperbolic model or the diffusive Kelvin-Voigt regime, depending on the magnitude of the viscoelastic parameters. In particular, the third residual reduces to
\begin{equation}
\begin{aligned}
&\mathcal{R}_3 = \hat{p} - F(\hat{A}), 
\quad &&\text{if } \tau_r \to 0 \text{ while } \eta \to 0 \\
& \quad &&\text{or } E_0 \to E_{\infty}, \\[1ex]   
&\mathcal{R}_3 = \hat{p} - F(\hat{A}) + \eta G(\hat{A}) \frac{\partial (\hat{A} \hat{u})}{\partial x}, 
\quad &&\text{if } \tau_r \to 0 \text{ while } E_0 \to \infty \\
& \quad &&\text{and } \eta \sim \tau_r E_0.
\end{aligned}
\end{equation}

{\color{black}
In the following numerical tests, the results obtained with the APNN described above are also compared against those of a vanilla PINN. In the standard PINN formulation, the governing equations are enforced through the conventional form of the viscoelastic constitutive relation. In this case, the first two residuals are defined as in the APNN formulation~\eqref{eq:apnn_res}, whereas the third residual is computed as
\begin{equation}
\mathcal{R}_{3}^{PINN} =
\biggl(
\frac{\partial \hat{p}}{\partial t}
+ E_0 G(\hat{A})\frac{\partial (\hat{A}\hat{u})}{\partial x}
\biggr)
+ \frac{1}{\tau_r}\bigl(\hat{p}-F(\hat{A})\bigr).
\end{equation}
}

\begin{remark}
It is important to note that due to the presence of variables in system~\eqref{eq:visc_model} that have very different scales, all equations used in this APNN framework must be expressed in their dimensionless form. This normalization is essential for stable and efficient network training, as it prevents variables with large differences in magnitude from dominating the gradients and ensures that all terms contribute properly to the loss function. For details on the derivation of the dimensionless form, we refer to~\cite{bertaglia2023multiscale}.
\end{remark}

\subsection{APNN implementation and validation datasets}\label{sec_dat}

To validate the proposed methodology, we present two types of numerical tests: one test case designed using a synthetic dataset that reproduces the average dynamics of the upper thoracic aorta (TA), computed by solving with an appropriate asymptotic-preserving Implicit-Explicit (IMEX) Runge-Kutta numerical scheme \cite{IMEXbook} system~\eqref{eq:visc_model}; three test cases performed using measured data from the right common carotid artery (CCA) of three healthy subjects.

In the following paragraphs, we discuss network design and training strategy of the APNN, details of the datasets, and of the preprocessing strategies.

\subsubsection{APNN architecture and training parameters}

In all the numerical tests performed, the APNN architecture comprises an input layer, three hidden layers of 32 neurons each, and an output layer. Hyperbolic tangent activation functions are used in all hidden layers. To enforce the physical positivity of the cross-sectional area and guarantee well-defined PDE residuals, the area output is transformed via a softplus activation function, ensuring $\hat{A}>0$. In contrast, linear activations are applied to the velocity and internal pressure outputs.

The network parameters $\boldsymbol{\theta}$ and the viscoelastic coefficients $\boldsymbol{\xi}=(\tau_r,E_0)$ are optimized simultaneously using the Adam optimizer~\cite{adam2014method}, with a learning rate $\lambda = 10^{-3}$. \textcolor{black}{Training is performed using a full-batch strategy, with an early stopping criterion applied based on the stability of the identified viscoelastic parameters $\boldsymbol{\xi}$. Specifically, training is terminated when the relative variation of both $E_0$ and $\tau_r$ does not exceed $1\%$ over a window of $10^6$ consecutive epochs. The maximum number of epochs is set to $12\cdot10^6$ for the synthetic dataset and $30\cdot10^6$ for the measured CCA datasets. These early stopping parameters and maximum number of epochs were chosen to ensure that the identified viscoelastic parameters $\boldsymbol{\xi}$ reached stable, approximately constant values during training. Computations were performed on a workstation equipped with an Intel Core i9-10920X CPU, 16~GB of RAM, and an NVIDIA Quadro P620 GPU with 2~GB of memory. The resulting average computational time per epoch is approximately $0.0121~\mathrm{s}$ for the synthetic dataset and $0.020~\mathrm{s}$ for the measured CCA datasets.}

%Training is performed using a full-batch strategy for $9\cdot10^6$ epochs for the synthetic dataset and $12\cdot10^6$ epochs for the measured CCA datasets, with no early stopping criterion applied. These numbers of epochs were chosen to ensure that the identified viscoelastic parameters $\boldsymbol{\xi}$ reached stable, approximately constant values during training. Computations were performed on a workstation equipped with an Intel Core i9-10920X CPU, 16~GB of RAM, and an NVIDIA Quadro P620 GPU with 2~GB of memory. The resulting average computational time per epoch is approximately $0.0121~\mathrm{s}$ for the synthetic dataset and $0.020~\mathrm{s}$ for the measured CCA datasets.}
To enforce physical positivity and enhance numerical stability, the viscoelastic parameters $\boldsymbol{\xi}$ are transformed via an exponential function before computing the physics-based constraints. This approach ensures that $\tau_r$ and $E_0$ remain strictly positive and allows the optimizer to operate more effectively across different orders of magnitude.

The supervised data loss $\mathcal{L}_d$~\eqref{eq:loss_d} is computed using measurements extracted at a single spatial location $x^d=x_m$ along the vessel. For the synthetic dataset, $x_m$ is chosen at the vessel midpoint. For the measured CCA dataset, $x_m$ corresponds to the axial location where clinical measurements were obtained, which coincides with the vessel midpoint for all subjects in the dataset. 
The corresponding temporal signals of cross-sectional area and axial velocity are evaluated at $N_d^{(t)}$ uniformly distributed time instants, where $N_d^{(t)}$ represents the total number of available time steps. \textcolor{black}{No pressure measurements are included in the supervised loss term, except for the equilibrium pressure $p_0$, which is provided as a known input parameter.}

%The corresponding temporal signals of cross-sectional area and axial velocity are sampled at $N_d^{(t)}$ uniformly distributed time instants. For the synthetic dataset, we set $N_d^{(t)}=120$; for the measured CCA dataset, $N_d^{(t)}$ corresponds to the total number of available time steps. No pressure measurements are included in the supervised term. 

The physics-based collocation points used to evaluate the residual loss $\mathcal{L}_r$~\eqref{eq:loss_r} are generated on a structured spatio-temporal grid. Spatial coordinates are sampled uniformly at all available vessel locations, while temporal coordinates are sampled at $N_r^{(t)}$ uniformly distributed time instants. For both the synthetic and the measured CCA datasets, $N_r^{(t)}$ corresponds to the total number of available time steps.

The initial conditions are enforced at the spatial location $x^i=x_m$ at the initial time $t=0$ using the diastolic values of area $A_0$ and pressure $p_0$ as defined in~\eqref{eq:loss_b}. Positivity constraints on the internal pressure are applied at the same collocation points used for $\mathcal{L}_r$, uniformly covering the spatio-temporal domain.

The total loss function is defined as in~\eqref{eq:loss_pinn}, with weights set to $\omega_d = 10$, $\omega_r = 1$, and $\omega_b = 1$. These values were empirically selected to emphasize the data loss while still accounting for the physics residuals, and were fixed for all experiments.

\textcolor{black}{We compare the APNN against a vanilla PINN on the synthetic TA dataset. Both models use the exact same network architecture, training parameters, and collocation points, differing only in the formulation of the $\mathcal{R}_3$ residual used to compute the physics-based loss.}

%\begin{remark}
%It is worth to highlight that the choice of a uniform spatio-temporal grid for the evaluation of the physics loss residuals has been made after testing different sampling modalities. Indeed, unexpectedly, for the test cases considered in this work, the uniform grid resulted to be the most efficient when compared to a random sampling and even to a resampling strategy for uniform sampling, as proposed in \cite{wu2023}.
%\textcolor{blue}{Amplia pure la trattazione qui dell'argomento.}
%\end{remark}

\begin{remark}
It is worth highlighting that the choice of a uniform spatio-temporal grid for the evaluation of the physics loss residuals has been made after testing different sampling modalities. In particular, we considered: (i) uniform grid sampling, (ii) fixed random sampling, (iii) random sampling with resampling at each epoch, and (iv) Residual-based Adaptive Distribution (RAD), where collocation points are iteratively redistributed in regions characterized by larger residuals, as proposed in \cite{wu2023}.
For the test cases considered in this work, all the sampling approaches provided comparable results in terms of both prediction accuracy and loss convergence rate. Therefore, the selection of the uniform grid was primarily motivated by computational efficiency, as fixed collocation points avoid the additional cost associated with repeated sampling, thereby reducing the cost per epoch. A similar argument applies to fixed random sampling; however, since it did not provide any clear advantage over the grid-based approach, the latter was ultimately preferred.
\end{remark}

\begin{remark}
\textcolor{black}{In the in vivo test cases, the optimization problem is more challenging than in the synthetic case, requiring an additional physical constraint. Specifically, from the definition of the relaxation time in~\eqref{eq:taur}, the condition $E_0 \geq E_\infty$ must be satisfied to ensure $\tau_r \geq 0$. Without explicitly enforcing this constraint, the optimizer may yield $E_0 < E_\infty$, resulting in a negative relaxation time, which is physically not meaningful within the SLS constitutive framework. To prevent this for the in vivo measured dataset, we optimize the difference $\Delta E = E_0 - E_\infty$ instead of $E_0$ directly. The positivity of $\Delta E$ is then guaranteed by applying the same logarithmic reparametrization adopted for $\tau_r$ and $E_0$, ensuring that the constraint is respected throughout the entire training process.}
\end{remark}

%\section{Datasets}\label{sec5}

\subsubsection{Synthetic dataset}\label{sec:synt_dat}
To generate synthetic data for the upper TA, the viscoelastic blood flow model~\eqref{eq:visc_model} is numerically solved. This multiscale hyperbolic system is integrated in time using a third-order AP IMEX Runge-Kutta scheme~\cite{boscarino2017unified}, which combines implicit and explicit time integration to efficiently handle the multiscale physical processes involved. In particular, the third equation of the system, where the stiff source term related to the vessel wall viscoelasticity appears, is implicitly integrated, while continuity and momentum equations are discretized explicitly. The spatial discretization is performed using a finite volume method and third-order WENO reconstructions \cite{shu1997}. Numerical fluxes and non-conservative jumps at cell interfaces are evaluated using the path-conservative Dumbser-Osher-Toro Riemann solver~\cite{leibinger2016path,dumbser2011,dumbser2011a}. The resulting scheme is AP, maintaining stability, accuracy and consistency even in the zero relaxation limit.

\begin{table}[b!]
\centering
\caption{Model parameters for the tapered upper TA used to generate the dataset. 
Geometrical and mechanical properties, taken from~\cite{xiao2014systematic}, include vessel length ($L_0$), inlet and outlet radii ($R_{0,\mathrm{in}}$, $R_{0,\mathrm{out}}$), wall thickness ($h_0$), reference wave speed ($c$), diastolic pressure ($p_0$), Windkessel parameters (resistances $R_1$, $R_2$, compliance $C$, outflow pressure $p_{\mathrm{out}}$), viscoelastic wall properties (instantaneous and asymptotic Young moduli $E_0$, $E_\infty$, viscosity $\eta$, relaxation time $\tau_r$), and blood density ($\rho$). 
Numerical setup: $n_c = 12$ cells corresponding to a uniform mesh size $\Delta x = 2.01~\mathrm{cm}$, and simulation time $t_\mathrm{m} = 20.0~\mathrm{s}$ (20 cardiac cycles, 63~bpm). 
The time step size $\Delta t$ is chosen as the minimum between the hyperbolic Courant--Friedrichs--Lewy (CFL) stability constraint $\Delta t \le \mathrm{CFL} \, \Delta x / \{\max |u \pm c|\}$ with $\mathrm{CFL} = 0.9$, and the parabolic stability restriction $\Delta t \le \nu \Delta x^2$ with $\nu = 0.5$~\cite{boscarino2017unified}.}
\begin{tabularx}{\columnwidth}{X c |X c}
\hline
\textbf{Parameter} & \textbf{Value} & \textbf{Parameter} & \textbf{Value} \\
\hline
$L_0$ [cm]             & 24.137   & $R_1$ [MPa s/m$^3$] & 18.503  \\
$R_{0,in}$ [mm]    & 15.000   & $R_2$ [MPa s/m$^3$] & 104.920 \\
$R_{0,out}$ [mm]   & 10.000   & C [m$^3$/GPa]          & 10.163  \\
$h_0$ [mm]         & 1.000    & $E_0$ [MPa]            & 0.727   \\
$c$ [m/s]          & 5.494    & $E_\infty$ [MPa]       & 0.533   \\
$p_0$ [kPa]        & 9.467    & $\eta$ [kPa s]         & 23.884  \\
$p_{out}$ [kPa]    & 0.000    & $\tau_r$ [s]           & 0.009   \\
$\rho$ [kg/m$^3$]      & 1060    \\
\hline
\end{tabularx}
\label{tab:tab1}
\end{table}

At the boundaries of the spatial domain, physiologically realistic conditions are imposed: the inflow is prescribed through a time-dependent inlet flow rate profile based on~\cite{boileau2015benchmark}, while the outflow is represented by a three-element Windkessel (RCR) model, accounting for peripheral resistance and compliance. The RCR elements are coupled to the 1D domain by solving the Riemann problem at the outlet interface.

The model parameters considered are listed in Table~\ref{tab:tab1}. The TA was modeled with a tapered geometry, with the radius varying linearly from the inlet ($R_{0,in}$) to the outlet ($R_{0,out}$). Outflow Windkessel parameters were calibrated following the approach in~\cite{alastruey2012physical, xiao2014systematic}, the viscoelastic parameters $E_0$ and $\tau_r$ were determined as in~\cite{bertaglia2020computational}, and $E_\infty$ was calibrated as described in Section~\ref{sec:calibration}. \textcolor{black}{It is worth noting that the viscoelastic parameters listed in Table~\ref{tab:tab1} place the synthetic TA dataset close to the hyperbolic asymptotic limit. Indeed, the relaxation time $\tau_r = 0.009~\mathrm{s}$ is small compared to the characteristic time scale of the cardiac cycle ($\sim1~\mathrm{s}$), and the two elastic moduli are relatively close to each other ($E_0 = 0.727~\mathrm{MPa}$, $E_\infty = 0.533~\mathrm{MPa}$), consistently with the condition $E_0 \to E_\infty$ that characterizes the hyperbolic scaling (see Section~\ref{sec:2.1.2.1}).}

For a detailed description of the numerical method and the boundary conditions, we refer to~\cite{bertaglia2023multiscale}. For further details on IMEX schemes we invite the reader to refer to \cite{IMEXbook}.

{\color{black}
\paragraph{Effect of parameter-identification error on the forward solution}
To quantify the effect of the parameter-identification error on the forward model solution, an independent forward analysis is performed on the synthetic TA test case after APNN training. Specifically, the parameters $\boldsymbol{\xi}=(\tau_r,E_0)$ identified by the APNN are used in the original 1D viscoelastic blood flow model~\eqref{eq:visc_model}, which is solved using the AP IMEX scheme described above. All other geometric and physical parameters, as well as the initial and boundary conditions, are kept unchanged. The resulting forward numerical solution is then compared with both the APNN predictions and the reference synthetic solution. Since the reference data are generated using the same 1D model and numerical framework, this comparison quantifies how the parameter-identification error propagates into the forward solution.
}

\subsubsection{In vivo measurements and data preprocessing}

In vivo measurements were performed on the right CCA of three healthy volunteers after obtaining informed consent. Arterial cross-sectional area and blood flow velocity were acquired using Doppler ultrasound imaging (Xario 100, Toshiba Medical Systems, Japan) equipped with a $4.8/11~\mathrm{MHz}$ linear transducer (Toshiba PLU-704BT)~\cite{oglat2018review}, while arterial pressure waveforms were measured using applanation tonometry (PulsePen, DiaTecne srl, Milan, Italy)~\cite{salvi2004validation}. All measurements were performed approximately at the midpoint of the vessel’s length.

\paragraph{Cross-sectional area}
B-mode ultrasound imaging was used to visualize a longitudinal vessel segment and quantify vessel geometry variations over the cardiac cycle. Ultrasound videos were acquired with a field of view of $36 \times 25~\mathrm{mm}$, a pixel resolution of $0.087 \times 0.087~\mathrm{mm}$, and a temporal resolution of $\Delta t_\text{B-mode} = 0.033~\mathrm{s}$ per frame (corresponding to a frame rate of $30~\mathrm{Hz}$).

The vessel cross-sectional area profile was automatically extracted through a multi-step preprocessing pipeline. Each frame was cropped to the region of interest, denoised via median filtering (disk radius = 3 pixels), and intensity-normalized using percentile-based contrast rescaling. The vessel boundaries were segmented using a Chan--Vese active contour approach~\cite{chan1999active}  ($\mu = 0.25$), with morphological operations (small object and hole removal) applied to refine the segmentation. The axial location of the velocity measurement site was dynamically tracked across all frames by automatically detecting the spatial coordinates of the Doppler sample volume marker. The marker position was used as a geometric reference to retain the two closest segmented vessel-wall regions defining the upper and lower boundaries of the vessel. The vessel mask was then reconstructed by inverting the wall segmentation. The vessel centerline was identified via medial axis transformation~\cite{lee1994building} and regularized to eliminate spurious branches. The local radius profile was obtained by extracting the Euclidean distance transform~\cite{maurer2003linear} values along the vessel centerline, where the distance transform computes the distance of each pixel from the nearest vessel boundary. The radius value corresponding to the axial location of the velocity measurement was extracted and the resulting radius waveform was smoothed using a 1D Gaussian filter ($\sigma = 2$). A representative cardiac cycle was then selected as the interval between two consecutive minima of the radius waveform. Finally, the cross-sectional area was calculated assuming a circular vessel geometry.

To estimate the vessel tapering, the radius profiles obtained from the entire B-mode recording were averaged over all frames, encompassing all cardiac cycles available in the recording, providing a time-averaged radius distribution along the visible vessel segment. To reduce segmentation inaccuracies near the image boundaries, the first and last 10\% of the visible vessel length were excluded. The time-averaged radii at the two ends of the remaining central segment were then used to define a linear tapering profile along the vessel. Using literature values for the right CCA length~\cite{choudhry2016vascular}, this linear profile was extrapolated to obtain the full-vessel inlet and outlet radii, $R_{0,\mathrm{in}}$ and $R_{0,\mathrm{out}}$, from which the corresponding cross-sectional areas were calculated and used to define the equilibrium area $A_0(x)$.

%From the lumen segmentation, the time-averaged cross-sectional area over the cardiac cycle was also calculated at both the start and the end of the vessel segment captured in the video. Using literature values for the right CCA length~\cite{choudhry2016vascular}, the inlet and outlet cross-sectional areas of the full vessel were extrapolated by assuming a linear tapering of the vessel radius along its length, varying from $R_{0,in}$ at the inlet to $R_{0,out}$ at the outlet.

\paragraph{Blood flow velocity}
Blood flow velocity was measured using pulsed-wave Doppler at the center of the B-mode vessel segment. The Doppler sample volume was set to $1.0$-$1.5~\mathrm{mm}$, with a beam-to-flow angle of $60^{\circ}$, and the resulting velocity waveform was acquired at a sampling interval of $\Delta t_\text{Doppler} = 0.0055~\mathrm{s}$.

Since the Doppler data were initially recorded as video sequences, post-processing was performed to extract a one-dimensional velocity waveform. An intensity threshold was applied to isolate the Doppler signal from the background, followed by a Laplacian-of-Gaussian filter to detect the signal boundaries. The maximum velocity envelope was then extracted by selecting the upper boundary of the Doppler signal over time and smoothed using a Savitzky--Golay filter. Points of maximum temporal derivative were subsequently identified to determine individual cardiac cycles. The extracted cycles were resampled to a common temporal grid and normalized to a standard duration. The aligned cycles were then averaged to obtain a representative velocity waveform.

Although the cross-sectional area and velocity signals are intrinsically synchronous, as they are acquired simultaneously by the Doppler ultrasound system, the independent extraction of a representative area cycle and the averaging of multiple velocity cycles may introduce a phase mismatch between the two waveforms. To correct for this discrepancy, the signals obtained over the entire recording duration were used to compute the temporal shifts between corresponding systolic peaks of the area and velocity waveforms for each cardiac cycle. The resulting shifts were averaged, and the mean temporal correction was applied to the averaged velocity waveform to ensure phase alignment with the cross-sectional area signal.

Finally, since Doppler measurements provide the centerline (maximum) velocity, a scaling factor of 0.5 was applied under the assumption of a parabolic velocity profile to estimate the mean cross-sectional velocity~\cite{quarteroni2004mathematical}.

\begin{table}%[b]
\centering
\caption{Vessel properties for the three studied arteries: CCA-A, CCA-B, and CCA-C. 
Geometrical and mechanical properties include age (descriptive variable, not included in the loss computation), vessel length ($L_0$), inlet and outlet radii ($R_{0,in}$, $R_{0,out}$), wall thickness ($h_0$), reference wave speed ($c$), diastolic pressure ($p_0$), asymptotic Young modulus ($E_\infty$), and blood density ($\rho$).}
\label{tab:tab2}
\begin{tabularx}{\columnwidth}{Xcccc}
\hline
\textbf{Parameter} & \textbf{CCA-A} & \textbf{CCA-B} & \textbf{CCA-C} \\
\hline
Age [years]          & 28 & 29 & 44 \\
$L_0$ [cm]               & 10.900 & 10.900 & 10.900\\
$R_{0,in}$ [mm]      & 3.267 & 3.514 & 3.653 \\
$R_{0,out}$ [mm]     & 2.809 & 2.899 & 2.332 \\
$h_0$ [mm]            & 0.300 & 0.300 & 0.300 \\
$c$ [m/s]     & 5.920 & 5.920 & 5.920 \\
$p_0$ [mmHg]         & 71.0 & 84.0 & 76.0 \\
$E_\infty$ [MPa]     & 0.752 & 0.794 & 0.741 \\
%$\eta$ [kPa s]       & 47.768 & 47.768 & 47.768 \\
$\rho$ [kg/m$^3$]    & 1060 & 1060 & 1060 \\
\hline
\end{tabularx}
\end{table}

\paragraph{Internal pressure}
Internal pressure waveforms were acquired using an applanation tonometry device consisting of a tonometric pen and an electrocardiography (ECG) unit. The pen was applied over the artery with moderate pressure to partially flatten the vessel wall, allowing intra-arterial pressure recording. The raw pressure signal was acquired at a sampling interval of $\Delta t_\text{Tonometer} = 0.001~\mathrm{s}$.

To reduce baseline variations in the ECG signal associated with slow respiratory oscillations, the ECG signal was processed by fitting a sixth-degree polynomial to estimate the low-frequency baseline component, which was subsequently subtracted from the raw ECG waveform. Individual cardiac cycles were identified by detecting the R-peaks of the ECG, which were used as temporal reference points to identify consecutive cardiac cycles in the pressure signal. The extracted pressure cycles were then resampled onto a common temporal grid and temporally normalized to a standard duration. Finally, the aligned cycles were averaged to obtain a representative pressure waveform.

Since the raw tonometric signal was acquired in arbitrary units, the resulting averaged waveform was subsequently calibrated into absolute pressure values ($\mathrm{mmHg}$) using brachial systolic and diastolic pressure measurements obtained via a sphygmomanometer at the beginning of the experiment. Specifically, a linear scaling was applied by matching the maximum and minimum values of the averaged tonometric waveform to the measured systolic and diastolic pressures, respectively. This calibration approach relies on the assumption that mean arterial pressure remains approximately constant from central to peripheral arteries, while diastolic pressure exhibits only minimal variations along the arterial tree~\cite{salvi2004validation}. Since pressure and cross-sectional area signals were acquired independently, the pressure cycles were temporally aligned with the area waveform by determining the shift that minimized the mean squared error between the averaged pressure waveform and the cross-sectional area waveform.\\

%The resulting pressure signal was then calibrated using brachial systolic and diastolic measurements obtained with a sphygmomanometer at the beginning of the experiment, providing absolute arterial pressure values. This approach is justified by the fact that mean arterial pressure is generally constant from the aorta to peripheral arteries, while diastolic pressure decreases only minimally from central to peripheral sites~\cite{salvi2004validation}. The resulting pressure waveform was temporally aligned with the cross-sectional area signal by applying the temporal shift that minimized the mean squared error between the averaged pressure cycle and the area waveform.\\

Finally, cross-sectional area, blood flow velocity, and internal pressure signals were resampled onto a uniform temporal grid to ensure consistent sampling across modalities. Their cardiac cycle duration was normalized to $T=1~\mathrm{s}$ and the resulting signals were then converted into their dimensionless form as described in Section~\ref{sec4}, allowing temporal comparability among modalities and subjects. It should be noted that, a single cardiac cycle was used for cross-sectional area due to its lower temporal resolution, which prevents reliable averaging without smoothing physiologically relevant features, such as the dicrotic notch. In contrast, velocity and pressure waveforms, having higher temporal resolution, were averaged over multiple cycles to reduce noise while preserving physiologically meaningful features. 

The geometric and mechanical properties of the three studied arteries (CCA-A, CCA-B, and CCA-C) are reported in Table~\ref{tab:tab2}.

\textcolor{black}{It is important to note that, although these preprocessing procedures are commonly adopted in the literature on hemodynamic data analysis~\cite{bertaglia2020computational}, the associated assumptions may affect the predictions of the proposed APNN framework and should therefore be taken into account when interpreting the results. These aspects will be addressed in future work through the integration of uncertainty quantification techniques into the proposed framework.}

\section{Results and Discussion}\label{sec5}

%To validate the proposed methodology, we present two types of numerical tests: one test case designed using a synthetic dataset that reproduces the average dynamics of the upper thoracic aorta (TA), computed by solving with an appropriate asymptotic-preserving Implicit-Explicit (IMEX) Runge-Kutta numerical scheme \cite{IMEXbook} system~\eqref{eq:visc_model}; three test cases performed using measured data from the right common carotid artery (CCA) of three healthy subjects.

%In the following sections, we discuss network design and training strategy of the APNN, details of the datasets, and the numerical results obtained.

\subsection{Synthetic data tests results}

\begin{figure*}%[h!]
    \centering
    \includegraphics[width=\textwidth]{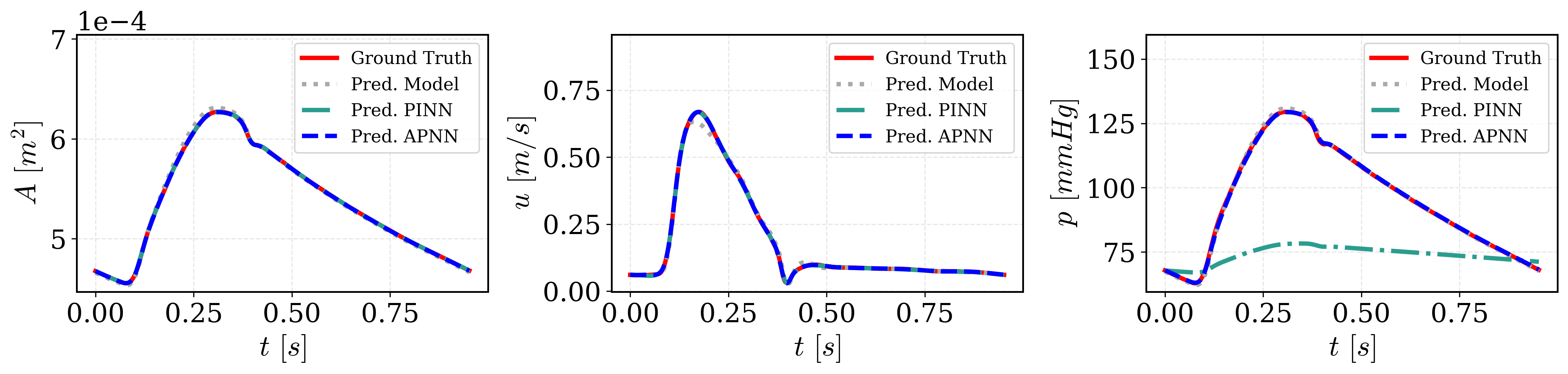}
    \\[2mm]
    \includegraphics[width=\textwidth]{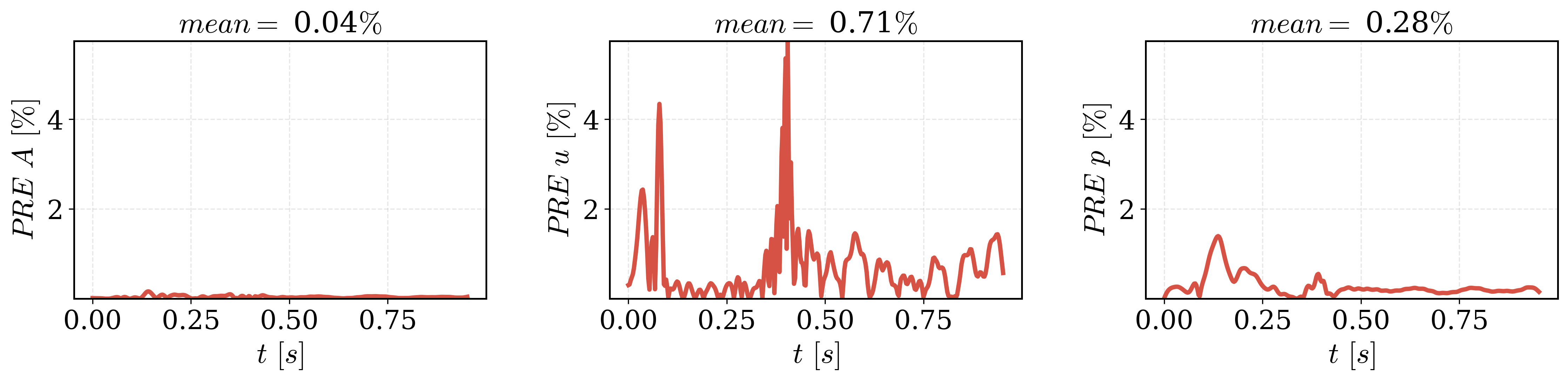}
\caption{Synthetic TA test results at the vessel midpoint for cross-sectional area, velocity, and pressure over one cardiac cycle. The first panel compares the reference numerical solution (red), the APNN reconstruction (blue), \textcolor{black}{the vanilla PINN reconstruction (green), and the solution obtained with the standard numerical solver (gray) using the parameters $E_0$ and $\tau_r$ inferred by the APNN}. The second panel shows the percentage relative error (PRE) \textcolor{black}{of the APNN reconstruction} over time, with the mean PRE reported for each variable. Area and velocity data sampled at this location are used for training, whereas pressure is inferred \textcolor{black}{entirely from the physics-based constraints.}}
\label{fig:TA}
\end{figure*}

\begin{table}
    \centering
    \color{black}
    \caption{\textcolor{black}{Mean percentage relative error (PRE) for the Synthetic TA dataset, computed over time at the midpoint and over the full spatial-temporal domain. Results are reported for the APNN, the vanilla PINN, and the forward solution of the 1D viscoelastic blood flow model~\eqref{eq:visc_model}, obtained using the parameters $E_0$ and $\tau_r$ inferred by the APNN and the numerical solver.}}
    \label{tab:mean_PRE}
    \begin{tabular}{lccc|ccc}
        \toprule
        & \multicolumn{3}{c|}{\textbf{Midpoint}} 
        & \multicolumn{3}{c}{\textbf{Full domain}} \\
        \textbf{Method} & $\boldsymbol{A}$ & $\boldsymbol{u}$ & $\boldsymbol{p}$ 
               & $\boldsymbol{A}$ & $\boldsymbol{u}$ & $\boldsymbol{p}$ \\
        \midrule
        APNN  & 0.04\% & 0.71\% & 0.28\%
              & 0.19\% & 1.09\% & 0.26\% \\
        PINN  & 0.03\% & 1.24\% & 21.12\%
              & 1.73\% & 2.09\% & 21.09\% \\
        Model & 0.37\% & 3.30\% & 0.59\%
              & 0.38\% & 7.91\% & 0.61\% \\
        \bottomrule
    \end{tabular}
\end{table}

As already discussed in Section \ref{sec4}, we perform the TA test case considering that both the instantaneous Young modulus $E_0$ and the relaxation time $\tau_r$ of the model are unknown and have to be estimated by solving the inverse problem. Moreover, we reconstruct the spatio-temporal dynamics of the state variables of the system. \textcolor{black}{To assess the performance of the proposed APNN, the results are directly compared against those obtained using a standard vanilla PINN under identical training conditions.}

The reconstruction at the vessel midpoint over a single cardiac cycle is shown in Figure~\ref{fig:TA} \textcolor{black}{and in Table~\ref{tab:mean_PRE}}. \textcolor{black}{Regarding the APNN,} the predicted waveforms for cross-sectional area and velocity closely follow the reference solution, indicating that the network accurately captures the local dynamics at the training location. The \textcolor{black}{APNN} inferred pressure also shows excellent agreement with the reference, despite the absence of direct pressure measurements in the training set. The corresponding mean percentage relative errors (PRE) are 0.04\% for area, 0.71\% for velocity, and 0.28\% for pressure. The largest discrepancies \textcolor{black}{in the APNN reconstruction} occur in the velocity signal near the systolic peak, where the local PRE reaches approximately 5\%, particularly during the systolic upstroke and early deceleration phase, when rapid temporal variations make the prediction more challenging.

\textcolor{black}{In contrast, while the vanilla PINN correctly captures the area and velocity profiles, it significantly underestimates the pressure signal, resulting in a curve with a visibly lower amplitude compared to the reference solution. This behavior is reflected in its mean PREs of 0.05\%, 1.24\%, and 21.12\% for area, velocity, and pressure, respectively, confirming the superior capability of the AP formulation in reconstructing unobserved variables.}

\begin{figure*}
    \centering
    \includegraphics[width=\textwidth]{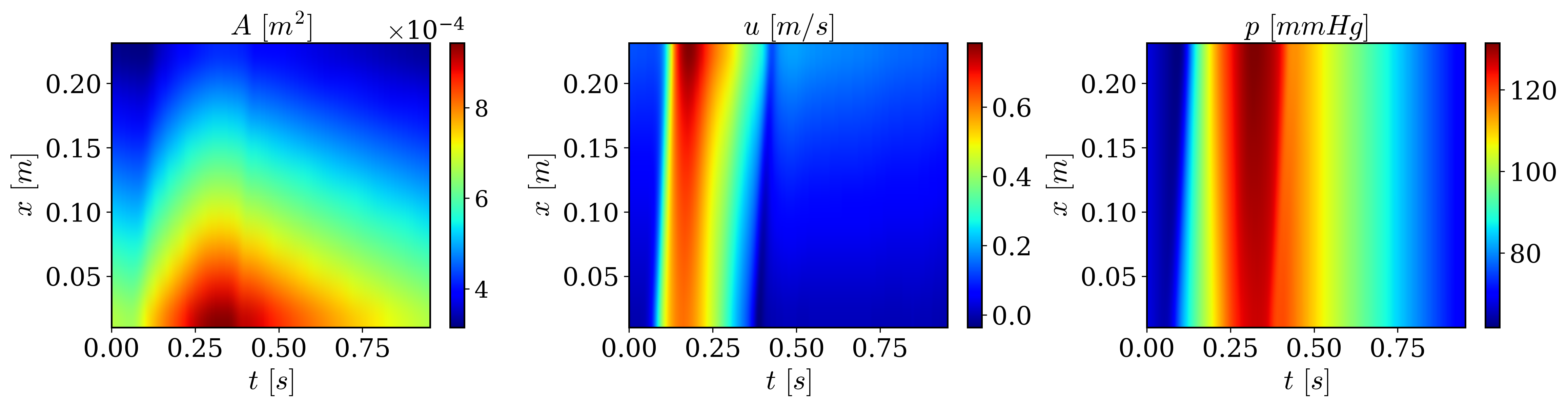}
    \\[2mm]
    \includegraphics[width=\textwidth]{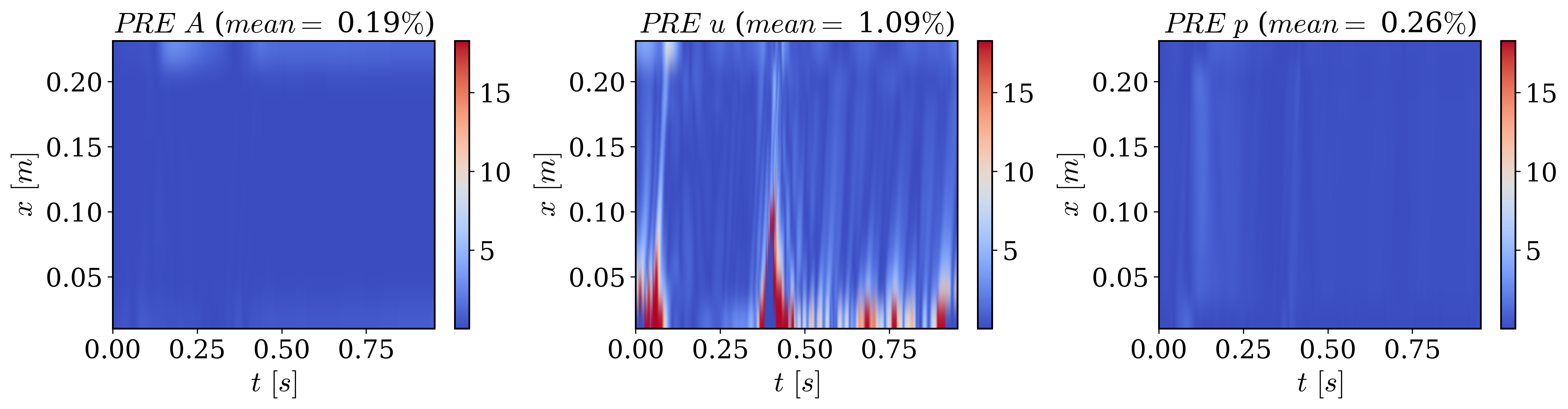}
\caption{Synthetic TA test results \textcolor{black}{for the APNN} over the full spatial domain. The first panel shows the APNN reconstruction of cross-sectional area, velocity, and pressure along the vessel over one cardiac cycle. The second panel displays the corresponding spatio-temporal percentage relative error (PRE) maps, with the mean PRE reported for each variable. Although training data for area and velocity were restricted to the vessel midpoint, the APNN provides the full-field distributions. The pressure field is entirely inferred by the APNN through the governing equations, without any direct measurements.}
\label{fig:TA_full}
\end{figure*}

The ability of the APNN to generalize over the full vessel domain is illustrated in Figure~\ref{fig:TA_full}, which shows the reconstructed spatio-temporal propagation of area, velocity, and pressure waves. The mean PREs over the full domain \textcolor{black}{for the APNN} are 0.19\% for area, 1.09\% for velocity, and 0.26\% for pressure \textcolor{black}{(Table~\ref{tab:mean_PRE})}. As observed at the midpoint, the largest discrepancies \textcolor{black}{in the APNN reconstruction} occur in the velocity signal near the systolic peak. These errors are more pronounced in the proximal region of the vessel, where they reach approximately 17\%, and progressively decrease toward the distal region, where they are approximately 5\%. This behavior is consistent with the expected waveform evolution along the vessel: the velocity profile exhibits a sharper systolic peak near the inlet, making the prediction more sensitive to rapid temporal variations, while it becomes smoother toward the outlet, leading to reduced errors. \textcolor{black}{In contrast, over the full spatial domain, the vanilla PINN yields mean PREs of 1.73\%, 2.09\%, and 21.09\% for area, velocity, and pressure, respectively (Table~\ref{tab:mean_PRE}).}

\begin{figure*}
    \centering
    \includegraphics[width=\textwidth]{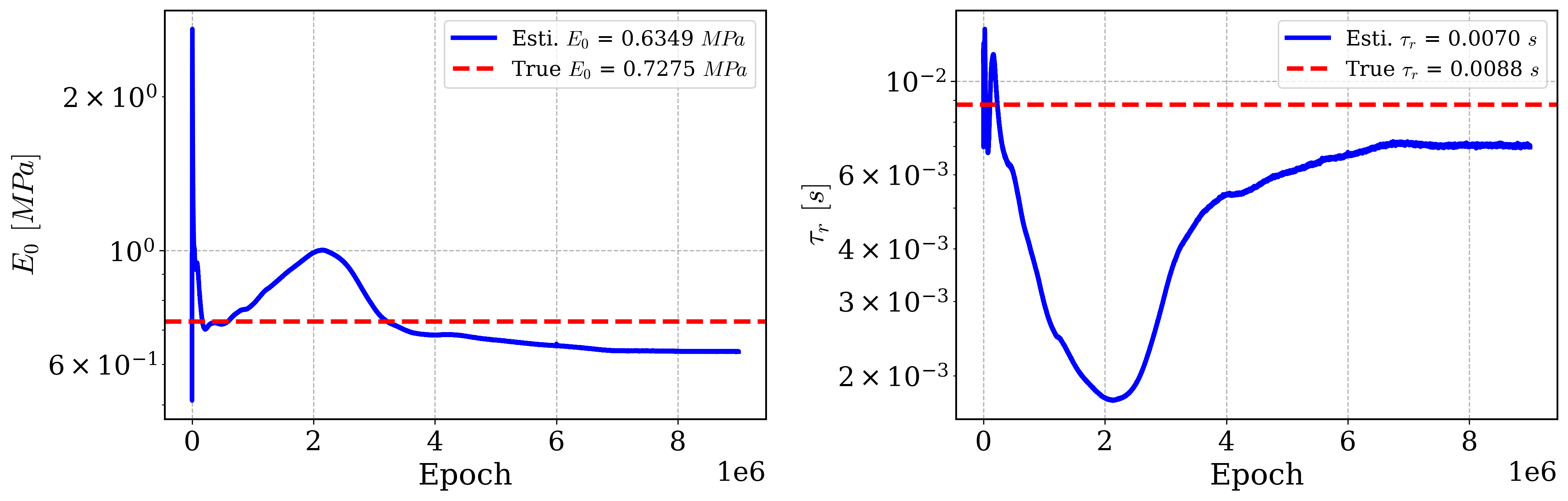}
\caption{Training progress of the APNN for the synthetic dataset. The evolution of the parameters $E_0$ and $\tau_r$ over the training epochs is shown in blue, while the corresponding ground truth values used to generate the data are shown in red. The legend reports the final estimated value of each parameter along with its ground truth.} %The mean Percentage Relative Error (PRE) for each variable is also indicated.}
\label{fig:TA_history}
\end{figure*}

The training evolution of the viscoelastic parameters $E_0$ and $\tau_r$ is reported in Figure~\ref{fig:TA_history} \textcolor{black}{for the AP configuration}. Both parameters progressively approach their reference values as the number of training epochs increases, converging to stable estimates. The final PREs \textcolor{black}{for the APNN} are 12.73\% for $E_0$ and 20.84\% for $\tau_r$. %While the errors on the viscoelastic parameters are moderate, they are fully compatible with the low prediction errors observed in the pressure curve, which is the primary quantity of clinical interest. These results indicate that the APNN can recover physiologically meaningful viscoelastic parameters while maintaining high accuracy in clinically relevant hemodynamic predictions.
These results suggest that the APNN can recover viscoelastic parameters that are in reasonable agreement with physiological expectations while maintaining high accuracy in clinically relevant hemodynamic predictions. \textcolor{black}{In contrast, the vanilla PINN fails to properly identify the viscoelastic parameters, converging to estimates with a PRE of $88.94\%$ for $E_0$ and $69667.73\%$ for $\tau_r$. This inaccurate estimation of the viscoelastic properties, particularly the relaxation time $\tau_r$, explains the poor reconstruction of the unobserved pressure field and highlights the benefits of the proposed AP formulation.}

\textcolor{black}{It is worth noting that the poorer performance of the vanilla PINN compared to the APNN — both in the reconstruction of the hemodynamic variables and in the identification of the viscoelastic parameters — occurs even in a configuration that lies close to the hyperbolic asymptotic limit (see Section~\ref{sec:synt_dat}). This finding suggests that even a configuration not too distant from the purely elastic regime can expose the limitations of standard PINN formulations, further motivating the adoption of an AP approach.}

{\color{black}
\paragraph{Effect of parameter-identification error on the forward solution}
To further quantify the impact of parameter-identification errors on the physical model solution, the parameters $E_0$ and $\tau_r$ inferred by the APNN were used to solve the original 1D viscoelastic model with the standard numerical solver. At the vessel midpoint, the resulting mean PREs were $0.37\%$, $3.30\%$, and $0.59\%$ for area, velocity, and pressure, respectively, whereas across the full spatial domain they reached $0.38\%$, $7.91\%$, and $0.61\%$, respectively. The forward numerical solution remains in reasonable agreement with the reference solution, suggesting that the identified parameters reproduce the main hemodynamic dynamics when propagated through the original governing equations. The APNN reconstruction achieves lower errors than the forward model for all variables, both at the vessel midpoint and across the full spatial domain. The lower errors observed for the APNN reconstruction may be attributed to the simultaneous optimization of the network weights and viscoelastic parameters, allowing the APNN to partially compensate for parameter-identification errors.}

\subsection{In vivo data tests results}
The proposed APNN framework was evaluated on three in vivo datasets of the CCA acquired from three subjects (CCA-A, CCA-B, CCA-C). In contrast to the synthetic case, ground truth information is available only at the measurement location, where cross-sectional area, velocity and pressure were recorded, while no reference solution is accessible over the full spatial domain. Furthermore, the viscoelastic parameters $E_0$ and $\tau_r$ used for comparison are derived from standard calibration procedures and should therefore be regarded as physiological estimates rather than exact ground truth values (which were clearly not available). %The model parameters employed to compute the APNN physics-based loss are reported in Table~\ref{tab:tab2}.

\begin{figure*}[p]
\centering
% -------- CCA-A --------
\subfloat[][]{%
  \begin{minipage}{\textwidth}
    \centering
    \textbf{CCA-A}\par
    \vspace{1mm}
    \includegraphics[height=0.138\textheight, keepaspectratio]{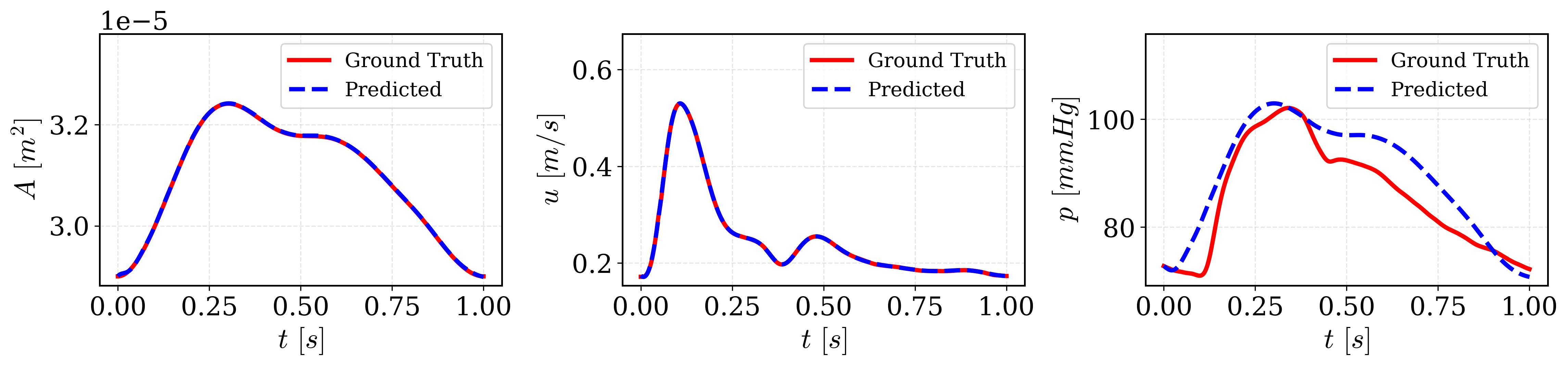}
    \includegraphics[height=0.138\textheight, keepaspectratio]{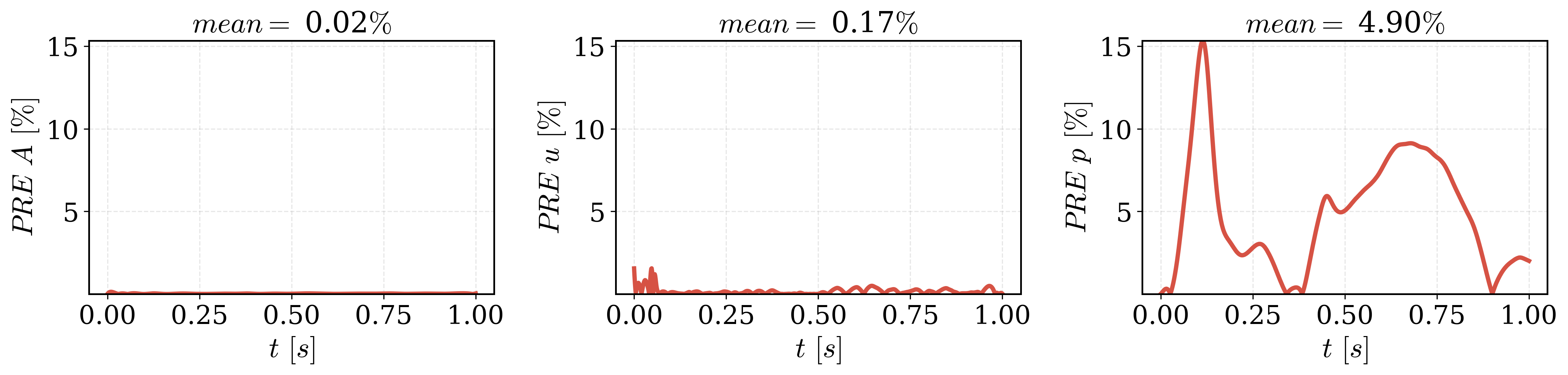}
  \end{minipage}%
}

%\vspace{0.5mm}
% -------- CCA-B --------
\subfloat[][]{%
  \begin{minipage}{\textwidth}
    \centering
    \textbf{CCA-B}\par
    \vspace{1mm}
    \includegraphics[height=0.138\textheight, keepaspectratio]{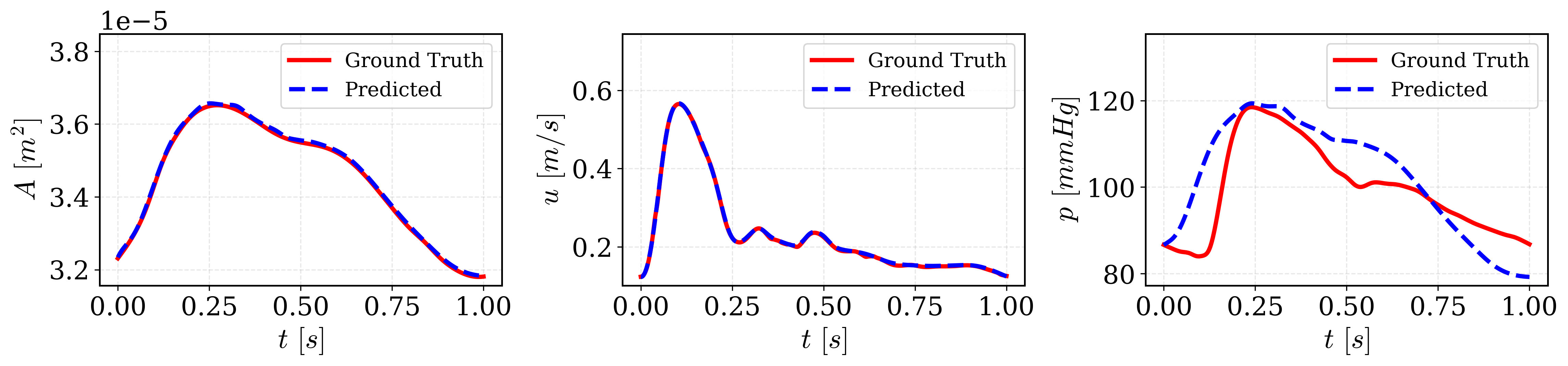}
    \includegraphics[height=0.138\textheight, keepaspectratio]{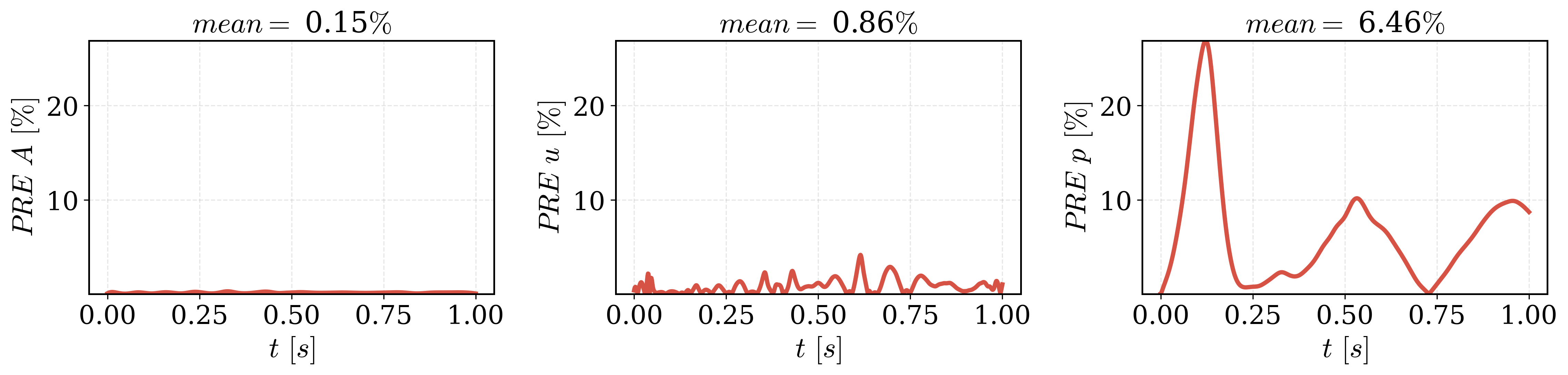}
  \end{minipage}%
}

%\vspace{0.5mm}
% -------- CCA-C --------
\subfloat[][]{%
  \begin{minipage}{\textwidth}
    \centering
    \textbf{CCA-C}\par
    \vspace{1mm}
    \includegraphics[height=0.138\textheight, keepaspectratio]{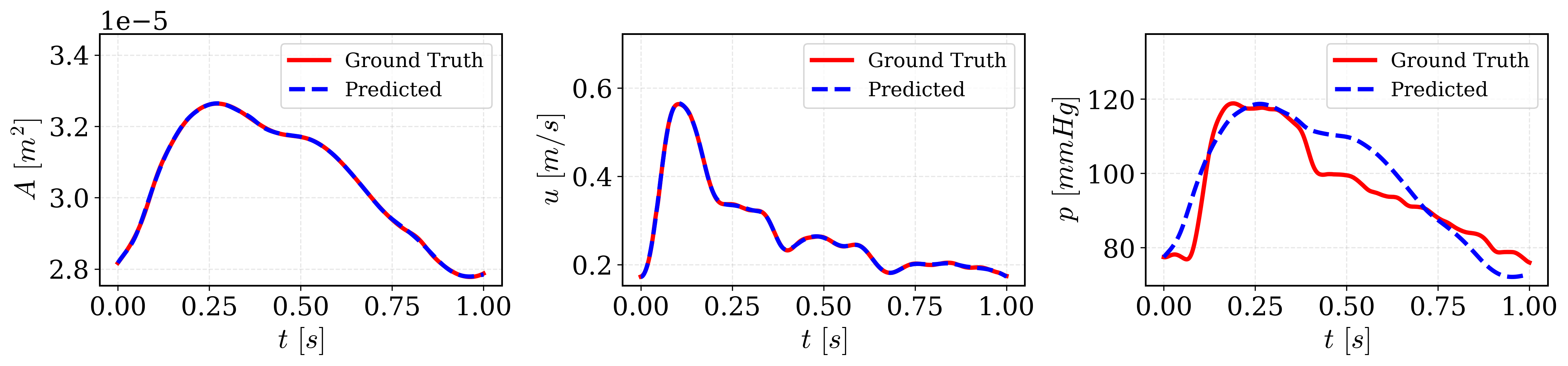}
    \includegraphics[height=0.138\textheight, keepaspectratio]{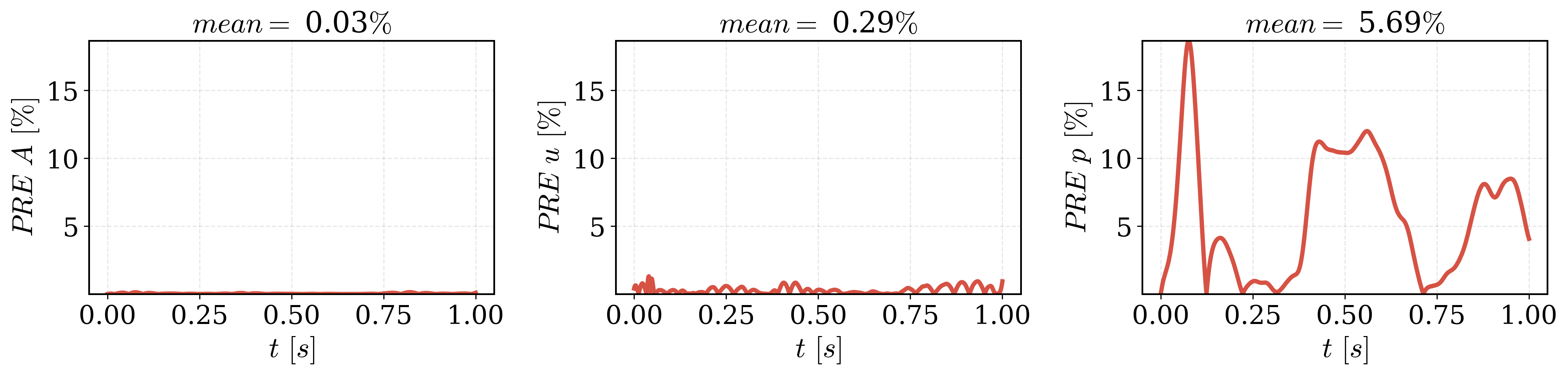}
  \end{minipage}%
}
\caption{In vivo test results at the vessel measurement point for three different carotid arteries (CCA-A, CCA-B, CCA-C). For each case, the upper panel shows the comparison between the measured waveform (red) and the APNN reconstruction (blue) for cross-sectional area, velocity, and pressure over one cardiac cycle. The lower panel reports the corresponding percentage relative error (PRE) over time, with the mean PRE indicated for each variable. Area and velocity measurements at the considered location are used for training, whereas pressure is entirely inferred by the APNN model.}
\label{fig:CCA}
\end{figure*}

Figure~\ref{fig:CCA} shows the reconstruction at the measurement point over one cardiac cycle. For all three cases, the APNN accurately reproduces the measured area and velocity waveforms, demonstrating its ability to capture subject-specific hemodynamics at the training location. The corresponding mean PRE values at the training point are 0.02\%, 0.15\%, and 0.03\% for area, and 0.17\%, 0.86\%, and 0.29\% for velocity in CCA-A, CCA-B, and CCA-C, respectively. The inferred pressure shows good overall agreement with the reference curve, despite the absence of direct pressure measurements in the training data. In particular, the systolic and diastolic values closely match the reference curve. The mean PRE values for pressure are 4.90\%, 6.46\%, and 5.71\% for CCA-A, CCA-B, and CCA-C, respectively. For pressure, slightly larger discrepancies are observed before the systolic peak, where the waveform exhibits sharper temporal gradients, and in correspondence with the dicrotic notch.

\begin{figure*}
    \centering
    % -------- CCA-A --------
    \subfloat[][]{%
        \begin{minipage}{\textwidth}
            \centering
            \textbf{CCA-A}\par
            \vspace{1mm}
            \includegraphics[width=\textwidth]{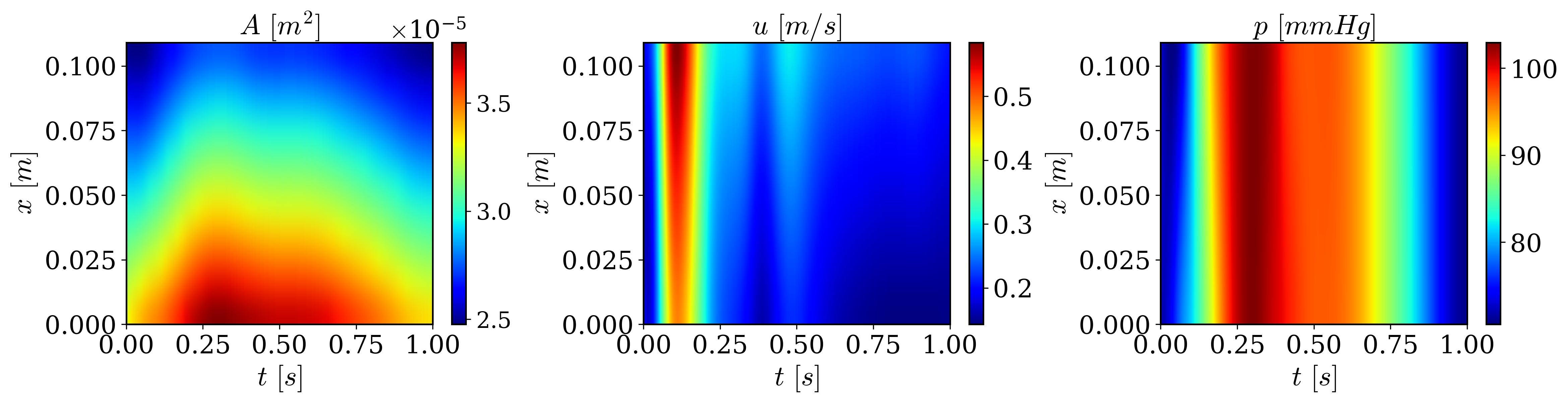}
        \end{minipage}%
    }
    
    % -------- CCA-B --------
    \subfloat[][]{%
        \begin{minipage}{\textwidth}
            \centering
            \textbf{CCA-B}\par
            \vspace{1mm}
            \includegraphics[width=\textwidth]{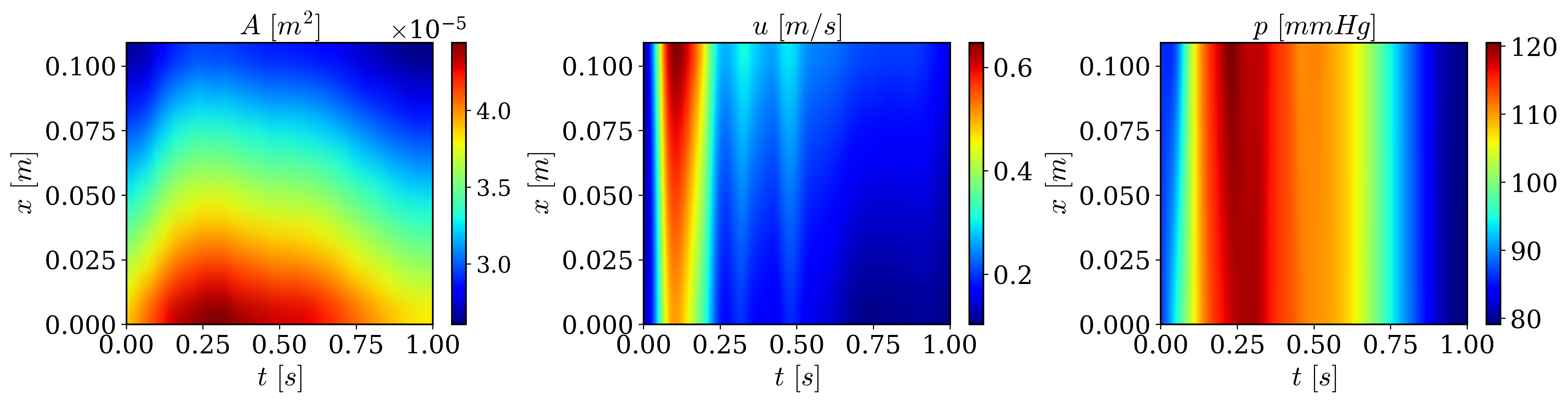}
        \end{minipage}%
    }
    
    % -------- CCA-C --------
    \subfloat[][]{%
        \begin{minipage}{\textwidth}
            \centering
            \textbf{CCA-C}\par
            \vspace{1mm}
            \includegraphics[width=\textwidth]{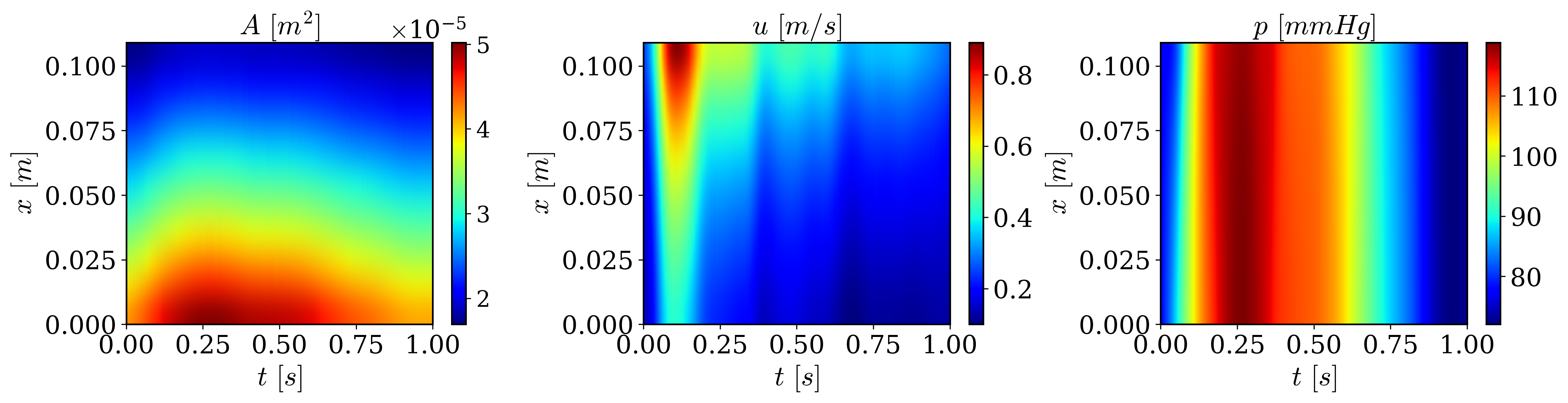}
        \end{minipage}%
    }
\caption{In vivo test results on the full spatial domain for three carotid arteries (CCA-A, CCA-B, CCA-C). For each case, the panels show the APNN reconstruction of cross-sectional area, velocity, and pressure over one cardiac cycle along the vessel. Measurements were available only at a single spatial location ($x_m$) for area and velocity, while the pressure field is entirely inferred by the APNN through the governing equations.}
\label{fig:CCA_2d}
\end{figure*}

Although the APNN was trained using area and velocity data from a single point along the vessel axis, it reconstructs the complete spatio-temporal fields of area, velocity, and pressure along the complete vessel length, as illustrated in Figure~\ref{fig:CCA_2d}. 
Since no complete ground truth is available in vivo, a quantitative error analysis over the entire domain is not possible. Nevertheless, the predicted distributions are physically plausible, showing a gradual decrease in area and a corresponding increase in velocity along the axial direction, consistent with expected hemodynamic behavior in arteries. 

\begin{figure*}[!t]
    \centering
    % -------- CCA-A --------
    \subfloat[][]{%
        \begin{minipage}{\textwidth}
            \centering
            \textbf{CCA-A}\par
            \vspace{1mm}
            \includegraphics[width=\textwidth]{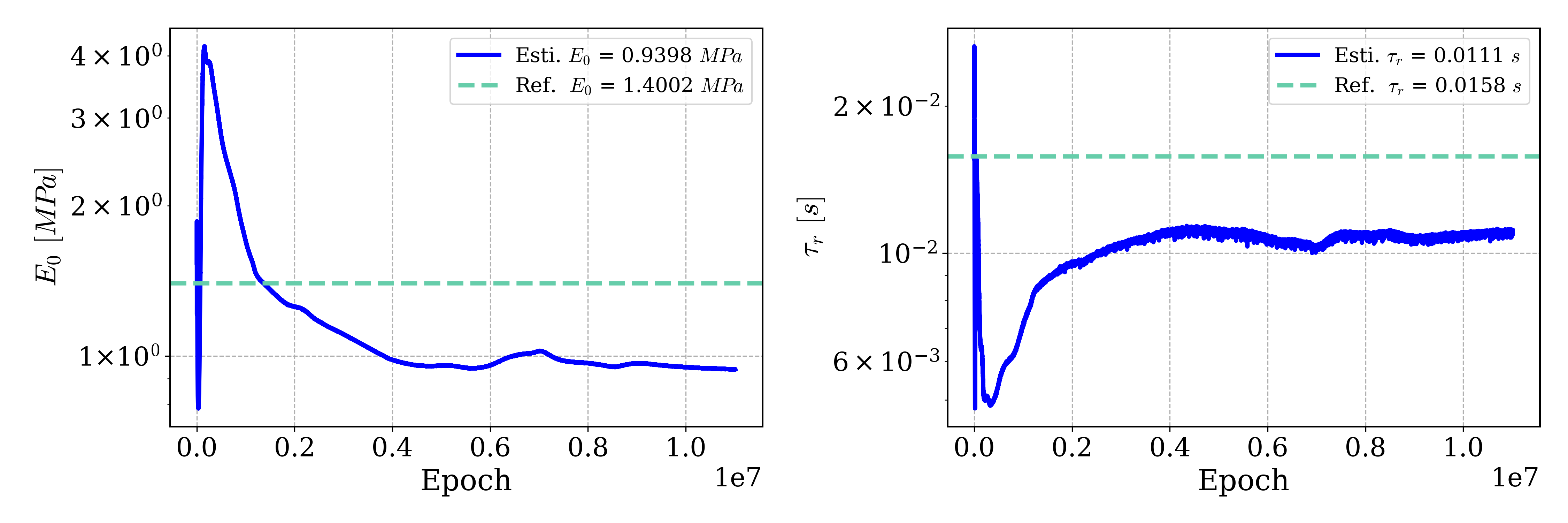}
        \end{minipage}%
    }
    
    % -------- CCA-B --------
    \subfloat[][]{%
        \begin{minipage}{\textwidth}
            \centering
            \textbf{CCA-B}\par
            \vspace{1mm}
            \includegraphics[width=\textwidth]{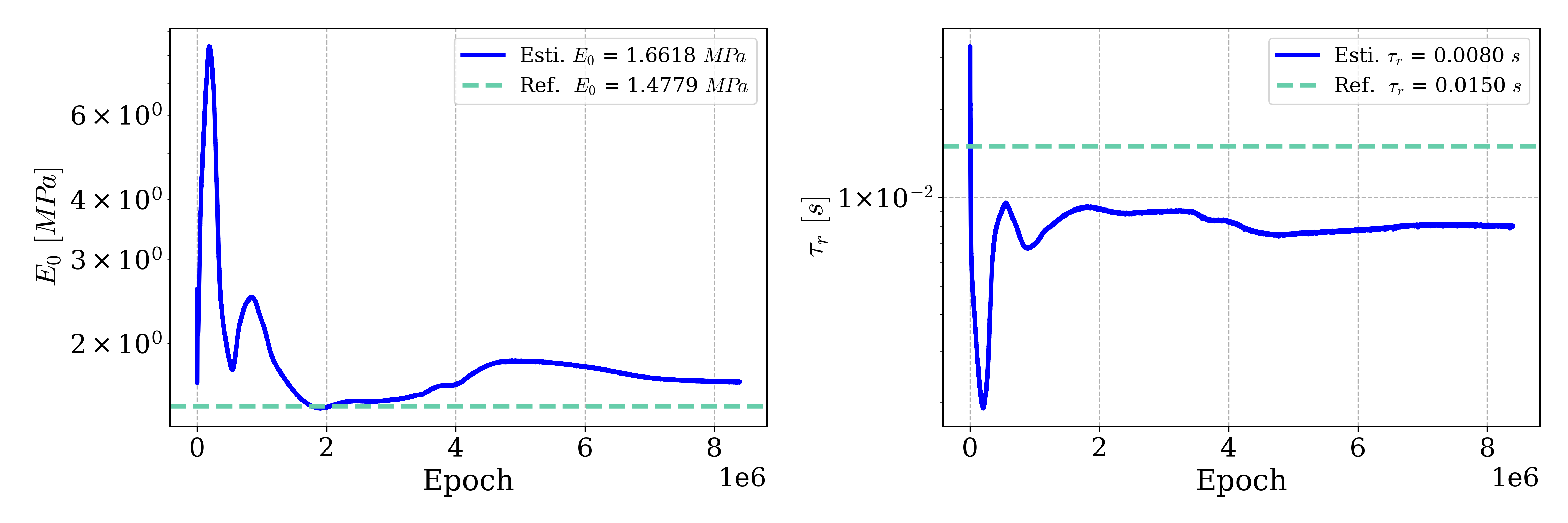}
        \end{minipage}%
    }
    
    % -------- CCA-C --------
    \subfloat[][]{%
        \begin{minipage}{\textwidth}
            \centering
            \textbf{CCA-C}\par
            \vspace{1mm}
            \includegraphics[width=\textwidth]{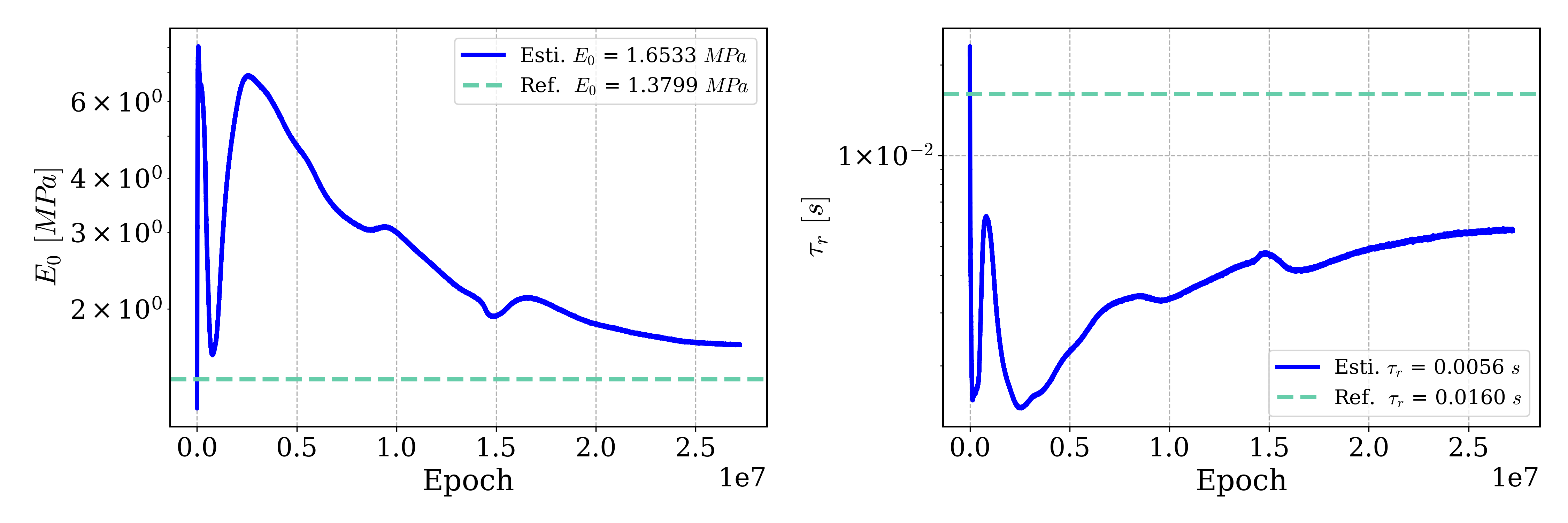}
        \end{minipage}%
    }
\caption{Training progress of the APNN for the three carotid arteries (CCA-A, CCA-B, CCA-C). The evolution of the parameters $E_0$ and $\tau_r$ over the training epochs is shown in blue, while reference values from standard calibration procedures are shown in green. These reference values are approximate estimates and do not represent exact ground truth. The legend reports the final estimated value of each parameter along with its corresponding reference value.}
\label{fig:CCA_history}
\end{figure*}

The training curves of $E_0$ and $\tau_r$ for the three arteries are reported in Figure~\ref{fig:CCA_history}. In all cases, both parameters converge toward stable values after an initial transient phase. The final estimates are in reasonable agreement and within the same order of magnitude as the reference values obtained from independent calibration procedures. However, it is important to emphasize that these reference values are themselves approximations and do not represent exact ground truth. Therefore, discrepancies between estimated and reference parameters cannot be interpreted strictly as identification errors, but rather as differences between two estimates derived under distinct assumptions.

%%==================================%%
%% CONCLUSIONS %%
%%==================================%%
\section{Conclusions}\label{sec6}

%In the present study, we applied a computational framework based on APNNs to address the challenge of non-invasively estimating intravascular blood pressure, typically accessible only in superficial vessels. 

\textcolor{black}{In the present study, we applied an APNN framework to the 1D multiscale viscoelastic blood flow model to address the challenge of non-invasively estimating intravascular blood pressure, typically accessible only in superficial vessels. By embedding the viscoelastic blood flow model within the learning procedure}, the APNN provides a physics-informed approach capable of reconstructing full pressure waveforms from readily available patient-specific data, such as cross-sectional area and velocity measurements via Doppler ultrasound. The framework also enables the identification of key viscoelastic parameters of the vessel wall (the instantaneous Young modulus and the relaxation time), which cannot be measured directly and whose conventional estimation methods are prone to errors. The AP formulation enclosed in the design of the neural network ensures consistency with the asymptotic hyperbolic and parabolic regimes of the underlying multiscale model, guaranteeing stable and physically coherent predictions across different scales of the viscoelastic parameters.

The framework was \textcolor{black}{assessed} on both synthetic and in vivo datasets. In both settings, the predicted pressure waveforms were consistent with expected values, despite pressure not being included in the supervised training data. Simultaneously, the identified viscoelastic parameters converged toward values \textcolor{black}{in line} with reference or independently calibrated estimates, \textcolor{black}{supporting the feasibility of the proposed inverse identification strategy.} Moreover, although the network was trained using cross-sectional area and velocity measurements at a single spatial location, it reconstructed \textcolor{black}{plausible} spatio-temporal evolution of the hemodynamic variables along the entire vessel length.

Overall, the results indicate that the APNN framework provides a \textcolor{black}{promising} and physically interpretable tool for non-invasive hemodynamic assessment. By combining patient-specific measurements with asymptotically consistent physics-informed modeling, the method enables the reconstruction of pressure and vessel wall properties with good agreement to reference data, while also delivering \textcolor{black}{reasonable} predictions in regions that are not directly accessible to measurement. \textcolor{black}{Taken together, these characteristics suggest that the proposed APNN could provide an alternative to conventional PDE-based formulations, as it can infer a plausible hemodynamic solution from sparse measurements without requiring the complete set of initial and boundary conditions necessary for a conventional forward numerical solution of the PDE system.}

Further studies could explore the use of alternative imaging modalities, such as magnetic resonance imaging~\cite{markl20124d, morgan20214d}, to achieve more detailed quantification of vessel and flow characteristics, as well as in vivo applications across different segments of the vascular network. These developments \textcolor{black}{could contribute to improving} the clinical applicability of the approach, enabling more comprehensive and patient-specific hemodynamic assessment.
Moreover, following \textcolor{black}{multi-fidelity, multi-level and multi-order approaches ~\cite{colebank2024,gao2020,bertaglia2026}}, we aim to further extend the proposed APNN framework by integrating it with uncertainty quantification techniques, in order to account for the inherent uncertainties associated with the measurement procedures used to collect biomedical data, \textcolor{black}{as well as those arising from preprocessing techniques and variability in APNN initialization, loss weights, and random seeds}.
\textcolor{black}{Finally, the theoretical analysis of well-posedness and practical identifiability for inverse problems solved through neural networks remains a challenging and active area of research. Investigating these aspects for the problem considered in this study therefore represents an important direction for future work.}

\subsection*{Acknowledgments}
This work has been written within the activities of the GNCS group of INdAM (Italian National Institute of High Mathematics), whose support is acknowledged. G.~B. has been funded by the European Union–NextGenerationEU, under the program “Future Artiﬁcial Intelligence—FAIR" (code PE0000013), MUR PNRR, Project “Advanced MATHematical methods for Artiﬁcial Intelligence—MATH4AI" and by MUR PRIN 2022 PNRR, Project No. P2022JC95T “Data-driven discovery and control of multiscale interacting artificial agent systems”.

\bibliography{refs}

\end{document}